\documentclass{article} % For LaTeX2e
\usepackage{graphicx} 
\usepackage{subcaption}
\usepackage{iclr2026_conference,times}
\usepackage[utf8]{inputenc}
\usepackage[T1]{fontenc}
\usepackage{hyperref}
\usepackage{url}
\usepackage{booktabs}
\usepackage{amsfonts}
\usepackage{nicefrac}
\usepackage{microtype}
\usepackage[table]{xcolor}
\usepackage{amsmath}
\usepackage{amssymb}
\usepackage{algorithm}
\usepackage{algorithmicx}
\usepackage{algpseudocode}
\usepackage{graphicx}
\usepackage{tikzducks}
\usepackage{multirow}
\usepackage{colortbl}
\usepackage{subcaption}
\usepackage{setspace}
\usepackage{enumitem}
\usepackage{mathtools}
\usepackage{lipsum}
\usepackage{xspace}
\usepackage{wrapfig}
\usepackage{makecell}
\usepackage{array}
\usepackage{caption}
\usepackage{ragged2e}
\usepackage{tabularx}
\usepackage{hyperref}
\usepackage{url}
\usepackage{footnote}

\usepackage{amsmath,amsfonts,bm}

\def\eqref#1{equation~\ref{#1}}
\def\1{\bm{1}}

\def\mI{{\bm{I}}}

\DeclareMathAlphabet{\mathsfit}{\encodingdefault}{\sfdefault}{m}{sl}
\SetMathAlphabet{\mathsfit}{bold}{\encodingdefault}{\sfdefault}{bx}{n}

\newcommand{\R}{\mathbb{R}}

\newcommand{\method}{FlashI2V\xspace}

\title{\textbf{\method}: \textbf{F}ourier-Guided \textbf{La}tent \textbf{Sh}ifting Prevents Conditional Image Leakage in Image-to-Video Generation}

\author{Yunyang Ge$^{1,3}$\qquad
Xinhua Cheng$^{1}$\qquad
Chengshu Zhao$^{1}$\qquad
Xianyi He$^{1,3}$\\
{\bf Shenghai Yuan$^{1}$}\qquad
{\bf Bin Lin$^{1,3}$}\qquad
{\bf Bin Zhu$^{1,3}$}\qquad
{\bf Li Yuan$^{1,2,\dagger}$\thanks{$^\dagger$Corresponding Author}} \\ 
$^1$Peking University, Shenzhen Graduate School\\
$^2$Peng Cheng Laboratory\\ 
$^3$Rabbitpre AI\\
\texttt{\{yunyang,chengxinhua,chengshuzhao,HeXianyi\}@stu.pku.edu.cn} \\
\texttt{\{yuanshenghai,linbin.ece,binzhu\}@stu.pku.edu.cn} \\
\texttt{yuanli-ece@pku.edu.cn}
}

\iclrfinalcopy % Uncomment for camera-ready version, but NOT for submission.
\makeatletter
\def\thanks#1{\protected@xdef\@thanks{\@thanks
        \protect\footnotetext{#1}}}
\makeatother

\begin{document}

\maketitle

\begin{abstract}
In Image-to-Video (I2V) generation, a video is created using an input image as the first-frame condition. Existing I2V methods concatenate the full information of the conditional image with noisy latents to achieve high fidelity. However, the denoisers in these methods tend to shortcut the conditional image, which is known as conditional image leakage, leading to performance degradation issues such as slow motion and color inconsistency. In this work, we further clarify that conditional image leakage leads to \textbf{overfitting} to in-domain data and decreases the performance in out-of-domain scenarios. Moreover, we introduce \textbf{F}ourier-Guided \textbf{La}tent \textbf{Sh}ifting \textbf{I2V}, named \textbf{\method}, to prevent conditional image leakage. Concretely, \method consists of: (1) \textbf{Latent Shifting.} We modify the source and target distributions of flow matching by subtracting the conditional image information from the noisy latents, thereby incorporating the condition implicitly. (2) \textbf{Fourier Guidance.} We use high-frequency magnitude features obtained by the Fourier Transform to accelerate convergence and enable the adjustment of detail levels in the generated video. Experimental results show that our method effectively overcomes conditional image leakage and achieves the best generalization and performance on out-of-domain data among various I2V paradigms. With only 1.3B parameters, \method achieves a dynamic degree score of 53.01 on Vbench-I2V, surpassing CogVideoX1.5-5B-I2V and Wan2.1-I2V-14B-480P. Project page: \url{https://pku-yuangroup.github.io/FlashI2V/}
\end{abstract}

\section{Introduction}
Conditional Video Generation~\citep{consisti2v,i2vgen_xl,animate_anyone,anyi2v,consisid,viewcrafter,fantasyid} refers to the technology that generates videos based on user-provided conditions, with significant applications being Text-to-Video (T2V) Generation and Image-to-Video (I2V) Generation. Since T2V generation produces a video solely based on a prompt, it struggles to accurately define scenes, such as accurate color and shape within the video. In contrast, I2V generation creates a video from both a user-provided image and a descriptive prompt, ensuring that the video content semantically aligns with the prompt and the first frame matches the provided image at the pixel level. In the commercial State-of-the-Art (SOTA) video generation product, Kling~\citep{kling}, 85\% of usage calls are for I2V generation.

\begin{figure}[t]
    \centering
    \begin{subfigure}[h]{0.768\textwidth}
        \centering
        \includegraphics[width=\textwidth]{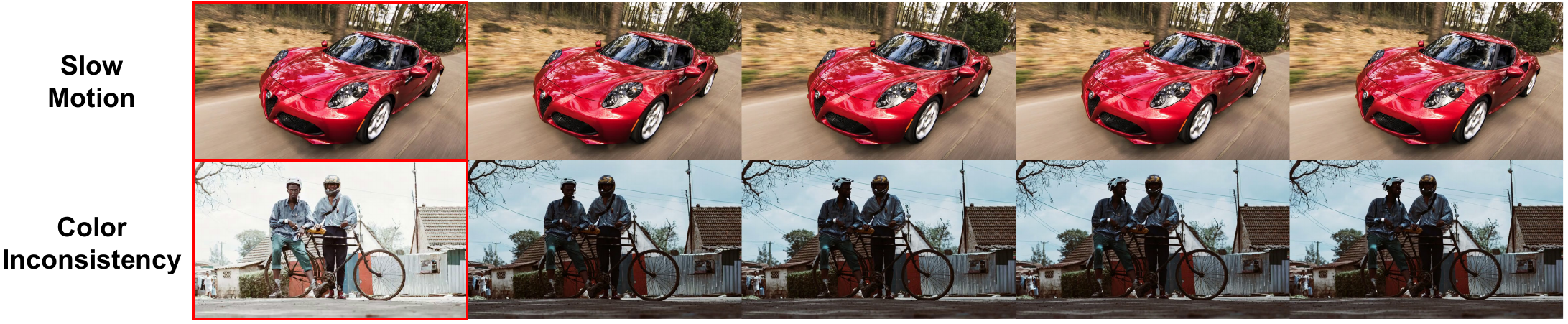}  
        \caption{Performance Degradation}
        \label{fig:introduction_inconsistency_qualitative}
    \end{subfigure}
    \begin{subfigure}[h]{0.212\textwidth}
        \centering
        \includegraphics[width=\textwidth]{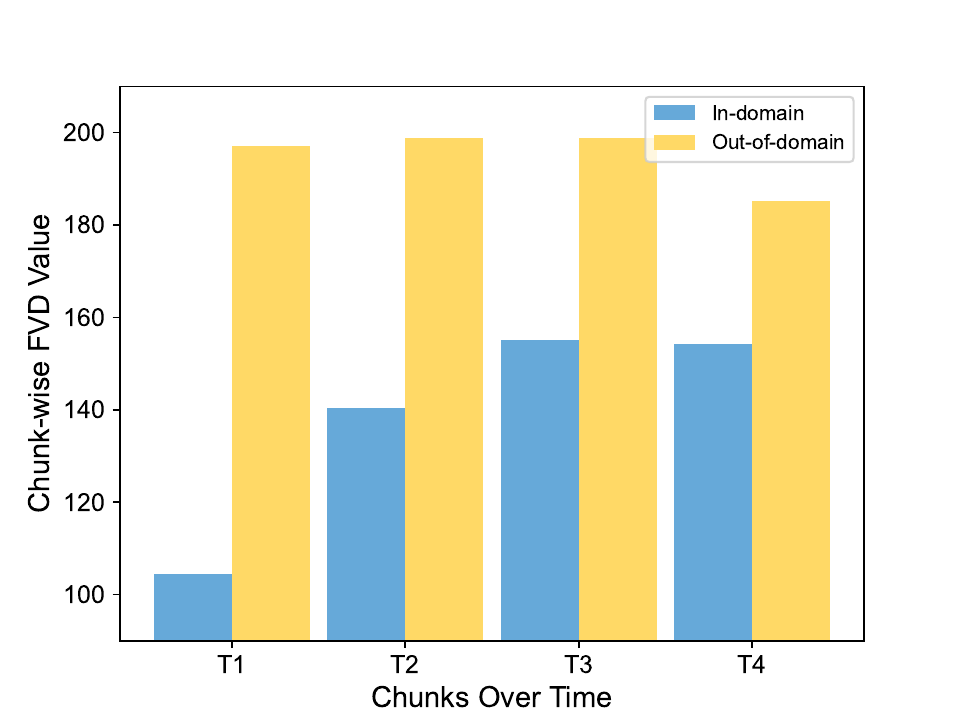}  
        \caption{Overfitting}
        \label{fig:introduction_inconsistency_quantitative}
    \end{subfigure}
    \caption{{\bf Conditional image leakage}. (a) Conditional image leakage causes performance degradation issues, where the videos are sampled from Wan2.1-I2V-14B-480P with Vbench-I2V text-image pairs. (b) In the existing I2V paradigm, we observe that chunk-wise FVD on in-domain data increases over time, while chunk-wise FVD on out-of-domain data remains consistently high, indicating that the law learned on in-domain data by the existing paradigm fails to generalize to out-of-domain data.}
    \label{fig:introduction_overfitting}
    \vspace{-0.5em}
\end{figure}

Leveraged in I2V methods including Stable Video Diffusion (SVD)~\citep{stable_video_diffusion}, Open-Sora Plan~\citep{open_sora_plan}, CogVideoX~\citep{cogvideox}, and Wan2.1~\citep{wan}, existing approaches concatenate the conditional image latents encoded by a Variational Autoencoder (VAE)~\citep{vae} with the noisy latents along the channel dimension, achieving exceptionally high fidelity for the first frame. However, previous works~\citep{conditional_image_leakage,ALG} highlight that existing methods suffer from conditional image leakage. Especially at large time steps, the denoiser directly utilizes the condition in a shortcut manner to minimize loss instead of performing the complex denoising process during training, resulting in slow motion in the generated output during inference. In addition to slow motion, we also observe other performance degradation issues such as color inconsistency in the generated video, as shown in Fig.~\ref{fig:introduction_inconsistency_qualitative}.

To investigate why conditional image leakage leads to performance degradation, we explore the generalization of the existing concatenating I2V paradigm. During training, the conditional image is the first frame of a video. In contrast, during inference, the conditional image can come from any source and is not necessarily the first frame of an existing video. The ability to generate reasonable and high-quality videos from any conditional image requires strong generalization in I2V methods. Since we cannot achieve the training dataset of any existing model, we train a model with weights initialized from Wan-T2V-1.3B using the existing concatenating I2V paradigm and compare its performance on both in-domain and out-of-domain data. Each video is divided into temporal chunks with an equal frame interval. We then compare the Fréchet Video Distance (FVD)~\citep{fvd} of the generated chunks with the ground truth chunks to assess the generation quality at different time points in the video. Theoretically, the first frame of the generated video must exactly match the conditional image, while subsequent frames lack such constraints, resulting in an increasing chunk-wise FVD over time. In a desired I2V paradigm, this increasing pattern should hold for both in-domain and out-of-domain data. As illustrated in Fig.~\ref{fig:introduction_inconsistency_quantitative}, experimental results reveal that chunk-wise FVD on in-domain data increases gradually over time. However, in out-of-domain data, chunk-wise FVD remains consistently high. By comparing the chunk-wise FVD variation patterns on in-domain and out-of-domain data, we conclude that even if the first frame matches the conditional image, shortcutting causes out-of-domain results to lack coherent video quality. The law learned from in-domain data fails to generalize to out-of-domain data, indicating that the concatenating paradigm faces an overfitting challenge, and a more reasonable paradigm is expected.

To prevent conditional image leakage, we propose a method that introduces conditions through \textbf{F}ourier-Guided \textbf{La}tent \textbf{Sh}ifting \textbf{I2V}, termed \textbf{\method}. The method consists of two parts: (1) \textbf{Latent Shifting}. Since flow matching imposes no restrictions on the source and target distributions, we encode the conditional image latents using a time-independent network and subtract the encoding from both the source and target distributions. The new velocity field is structurally the same as the velocity field in the original T2V model. The time-independent network is initialized to zero, ensuring that the input of the denoiser remains unchanged at the beginning of training. As a result, the denoiser gradually learns to utilize information from the conditional image through the shifted latents. Latent shifting requires recovering content from a mix of noisy latents and condition information. At larger time steps, the lower signal-to-noise ratio makes content recovery more difficult, which fundamentally prevents the leakage caused by shortcutting. (2) \textbf{Fourier Guidance}. Since the conditional image information needs to be recovered from the shifted latents, latent shifting requires more time and data to achieve first-frame fidelity comparable to existing methods. To accelerate convergence, we apply the Fourier Transform to extract high-frequency magnitude features from the conditional image latents and concatenate them with noisy latents. Since these magnitude high-frequency features only represent the relative strength of the signal, they serve as a supplement to latent shifting, which cannot lead to shortcutting. Moreover, by adjusting the cutoff frequency of the Fourier Transform, we can easily control the detail level in the generated video.

Compared to various existing I2V paradigms, only \method demonstrates the same FVD variation pattern on both in-domain and out-of-domain data, indicating that it avoids the leakage caused by shortcutting the conditional image. Furthermore, \method achieves the lowest FVD value among different I2V paradigms, showcasing its excellent performance. With only 1.3B parameters, \method achieves comparable scores to CogVideoX1.5-5B-I2V and Wan2.1-I2V-14B-480P on Vbench-I2V~\citep{vbench,vbench2} and obtains a dynamic degree score of 53.01, significantly outperforming the other two methods with larger parameter sizes.

In summary, our contributions are as follows: (1) By analyzing the chunk-wise FVD variation patterns in various existing I2V paradigms, we show that conditional image leakage causes overfitting to in-domain data, leading to performance degradation issues like slow motion and color inconsistency during inference. (2) We propose latent shifting, which implicitly introduces conditions based on flow matching characteristics. Additionally, we use high-frequency magnitude features from the Fourier Transform as guidance to accelerate convergence and enable the flexible control of detail levels in the generated video. (3) Experimental results show that \method exhibits the best generalization and performance across various I2V paradigms and effectively avoids overfitting caused by conditional image leakage. Specifically, with only 1.3B parameters, \method achieves a dynamic degree score of 53.01, surpassing CogVideoX1.5-5B-I2V and Wan2.1-I2V-14B-480P.

\section{Related Work}
\subsection{Text-to-Video Generation}
In recent years, Text-to-Video Generation has made significant progress. T2V methods often use diffusion models~\citep{ddpm,ddim,score_diffusion} to model the generation process. Previous works typically employ UNet~\citep{unet} and add temporal transformers after image weights (commonly referred to as the 2+1D paradigm) as denoisers~\citep{magictime,chronomagic,animatediff,lavie,videocrafter1,videocrafter2}. After the release of Sora~\citep{sora}, the community uses Diffusion Transformers (DiTs)~\citep{dit,lightingdit} as denoisers~\citep{open_sora,open_sora_plan,latte,easyanimate} within the 2+1D paradigm to achieve T2V. To overcome the limited capability of the 2+1D paradigm in temporal modeling, approaches like Open-Sora Plan v1.2~\citep{open_sora_plan} model all tokens uniformly (commonly referred to as the 3D paradigm) instead of differentiating between image and temporal weights. At present, with the adoption of 3D Transformer and more advanced diffusion models, flow matching~\citep{flow_matching,rectified_flow}, T2V models can generate highly realistic videos.
\subsection{Image-to-Video Generation}
Image-to-Video Generation leverages a conditional image and a prompt as inputs, enhancing the controllability of the generated video. Stable Video Diffusion (SVD)~\citep{stable_video_diffusion} combines conditional image latents with noisy latents, and injects high-level semantic information of the conditional image extracted by CLIP~\citep{clip,languagebind,lin2023video,lin2024moe,chen2024sharegpt4video} into the denoiser. DynamiCrafter~\citep{dynamicrafter} improves on SVD by using a query transformer~\citep{query_transformer} to extract CLIP tokens. CogVideoX~\citep{cogvideox} concatenates zero-padding conditional image latents with noisy latents to introduce conditions. SEINE~\citep{seine} introduces a temporal inpainting model, where conditional image latents and mask sequences are concatenated with noisy latents to fill in subsequent frames. Open-Sora Plan v1.3~\citep{open_sora_plan,li2025wf} further expands the inpainting model to cover more tasks and proposes a progressive training strategy to enhance performance. Wan2.1~\citep{wan} improves the inpainting model by introducing semantic information extracted by CLIP. All of these approaches inject full conditional image information into the denoiser through a concatenation operation, resulting in excellent fidelity for the first frame.
\subsection{Conditional Image Leakage}
Conditional image leakage~\citep{conditional_image_leakage, opens2v} is an issue where the model shortcuts the conditional image information, especially at large time steps, rather than utilizing it as an auxiliary to generate the video from noisy latents. SVD introduces adding a small amount of noise to the conditional image to increase the dynamic degree, marking the first attempt to reduce conditional image leakage. Previous work~\citep{conditional_image_leakage} proposes addressing the leakage by starting the generation process from an earlier time step during inference and designing a time-dependent noise distribution for the conditional image during training. Additionally, Adaptive Low-pass Guidance (ALG)~\citep{ALG} is a training-free approach by using the low-pass information of the conditional image rather than its full information at large time steps. At present, resolving conditional image leakage remains an open problem, with no universally accepted solution in the community.
\section{Method}
In this section, we first introduce the preliminary knowledge of flow matching in Sec.~\ref{method:preliminary}. Then, in Sec.~\ref{method:latent_shifting}, we present latent shifting for introducing conditions implicitly based on the characteristics of flow matching. Finally, in Sec.~\ref{method:fourier_guidance}, we bring in Fourier guidance that injects high-frequency magnitude features extracted from the Fourier Transform into the denoiser to serve as a supplement.
\subsection{Preliminary for Flow Matching}
\label{method:preliminary} 
Continuous Normalizing Flows (CNFs)~\citep{CNFs} aim to learn a transformation from a sample $\boldsymbol{z}_1$ from a source distribution $q_1(\boldsymbol{z})$ to a sample $\boldsymbol{z}_0$ from a target distribution $q_0(\boldsymbol{z})$, where $q_0(\boldsymbol{z})$ represents the data distribution, and $q_1(\boldsymbol{z})$ is typically a known prior distribution, such as a standard normal distribution. This transformation is usually modeled as an ordinary differential equation (ODE) with $t \in [0, 1]$. Let $\boldsymbol{z}_t$ represents the intermediate state from $\boldsymbol{z}_1$ to $\boldsymbol{z}_0$, then the transformation is governed by the following equation:
\begin{equation}
    \frac{d\boldsymbol{z}_t}{dt}=\boldsymbol{v}_t(\boldsymbol{z}_t,t), t \in [0, 1].
\end{equation}
Here, $\boldsymbol{v}_t(\boldsymbol{z}_t,t)$ defines the velocity field at any time, dictating how the distribution transfers over time. 
The concept of Flow Matching (FM)~\citep{flow_matching,rectified_flow,improving_flow_matching} is to directly learn the vector field $\boldsymbol{v}_t(\boldsymbol{z}_t,t)$ from 
$\boldsymbol{z}_1$ to $\boldsymbol{z}_0$ using a neural network $\boldsymbol{v}_\theta(\boldsymbol{z}_t,t)$. Specifically, for $\boldsymbol{z}_1 \sim q_1$ and $\boldsymbol{z}_0 \sim q_0$, their linear interpolation is constructed as follows:
\begin{equation}
    \boldsymbol{z}_t=(1-t)\boldsymbol{z}_0+t\boldsymbol{z}_1, t\in[0,1].
\end{equation}
The vector field $\boldsymbol{v}_t(\boldsymbol{z}_t,t)$ in this interpolation mode is given by:
\begin{equation}
    \boldsymbol{v}_t(\boldsymbol{z}_t,t) = \frac{d\boldsymbol{z}_t}{dt} = \boldsymbol{z}_1 - \boldsymbol{z}_0.
\end{equation}
$\boldsymbol{v}_t(\boldsymbol{z}_t,t)$ is only related to the two points $\boldsymbol{z}_0$ and $\boldsymbol{z}_1$ of the probability path and is independent of $t$. The optimization objective of FM is to train a neural network $\boldsymbol{v}_\theta(\boldsymbol{z}_t,t)$ to approximate $\boldsymbol{v}_t(\boldsymbol{z}_t,t)$ using the Mean Squared Error (MSE) loss. Under a condition $\boldsymbol{y}$, flow matching can be modeled as Conditional Flow Matching (CFM). The optimization objective of CFM is:
\begin{equation}
    \mathcal{L}_{\mathrm{CFM}}(\theta)=\mathbb{E}_{t\sim\mathcal{U}[0,1],\boldsymbol{z}_0\sim q_0,\boldsymbol{z}_1\sim q_1}\left[\left\|\boldsymbol{v}_\theta((1-t)\boldsymbol{z}_0+t\boldsymbol{z}_1,t,\boldsymbol{y})-(\boldsymbol{z}_1-\boldsymbol{z}_0)\right\|_2^2\right],
\end{equation}
where $\mathcal{U}[0,1]$ represents the uniform distribution over $[0,1]$. During sampling, we sample $\boldsymbol{z}_1 \sim q_1$ and solve the ODE $\frac{d\boldsymbol{z}_t}{dt}=\boldsymbol{v}_\theta(\boldsymbol{z}_t,t, \boldsymbol{y})$ step by step from $t=1$ to $t=0$ to obtain $\boldsymbol{z}_0 \sim q_0$. Unlike Denoising Diffusion Probabilistic Models (DDPM)~\citep{ddpm}, which require the source distribution to be a standard normal distribution, FM imposes no such constraints on the source distribution. This flexibility allows FM to be used for transferring between any two distributions.

\begin{figure}[t]
    \centering
    \includegraphics[width=\linewidth]{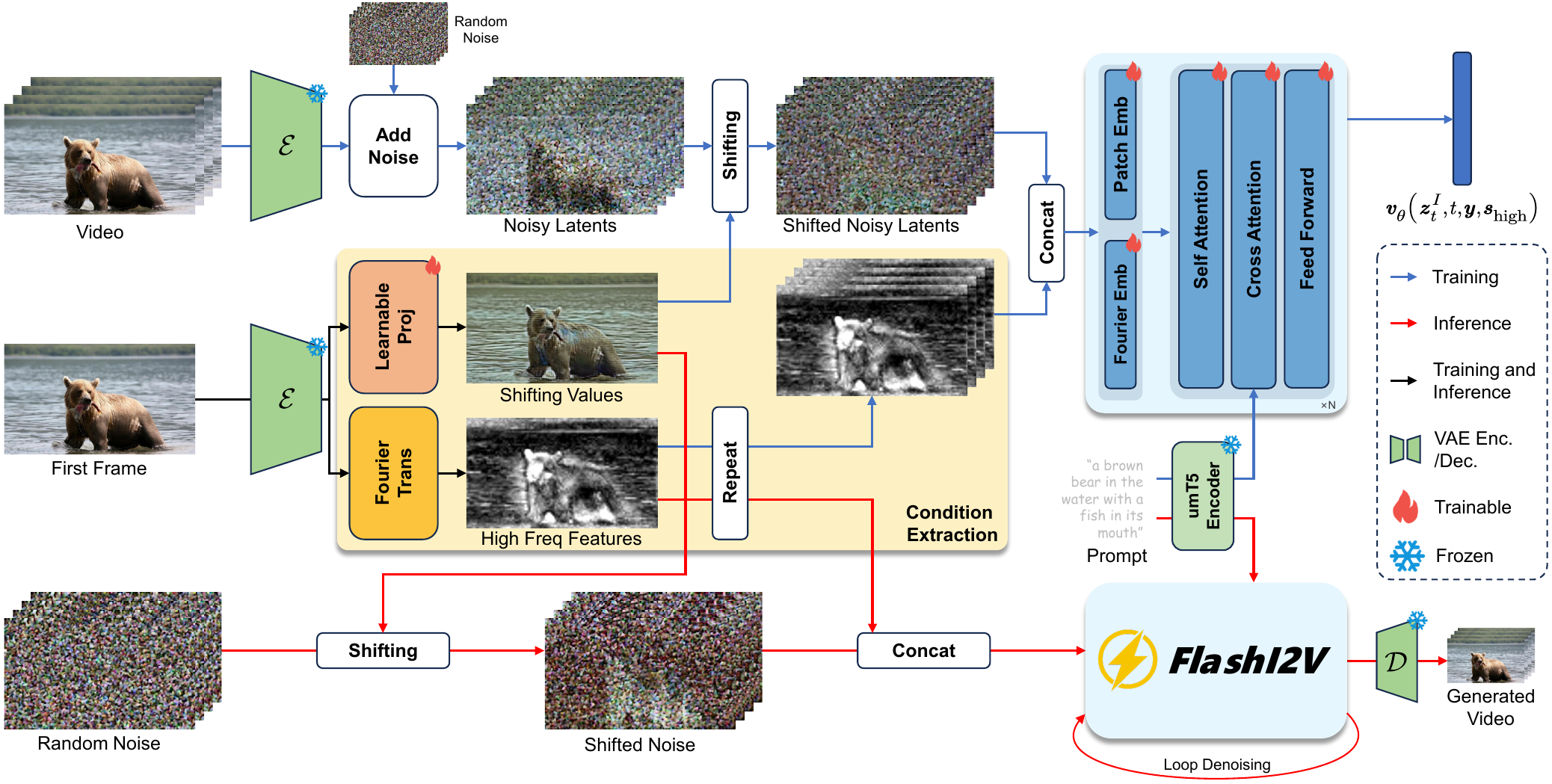}
    \caption{{\bf Method overview}. We extract features from the conditional image latents using a learnable projection, followed by the latent shifting to obtain a renewed intermediate state that implicitly contains the condition. Simultaneously, the conditional image latents undergo the Fourier Transform to extract high-frequency magnitude features as guidance, which are concatenated with noisy latents and injected into DiT. During inference, we begin with the shifted noise and progressively denoise following the ODE, ultimately decoding the video.}
    \label{fig:method_overall}
    \vspace{-1em}
\end{figure}

\subsection{Latent Shifting}
\label{method:latent_shifting}
We consider implementing I2V without explicitly incorporating the full information of the conditional image into the hidden states of the denoiser. Let $\boldsymbol{\epsilon} \sim \mathcal{N}(\mathbf{0},\mI)$, where $\mathcal{N}(\mathbf{0},\mI)$ denotes a standard normal distribution. Let a conditional image be $\boldsymbol{S} \in \R^{c \times h \times w}$, and a video starting with the conditional image be $\boldsymbol{X} \in \R^{c \times t \times h \times w}$, which means $\boldsymbol{X}[:,0] = \boldsymbol{S}$. Let $\mathcal{E}$ represent the encoder of the Variational Autoencoder (VAE)~\citep{vae}. Denote the source distribution sample as $\boldsymbol{z}_1^{T}$ and the target distribution sample as $\boldsymbol{z}_0^{T}$ for the T2V task, we have $\boldsymbol{z}_1^T=\boldsymbol{\epsilon}$, $\boldsymbol{z}_0^T=\boldsymbol{x}=\mathcal{E}(\boldsymbol{X})$, and the intermediate state $\boldsymbol{z}_t^{T}$ at any time $t$ under FM is given by:
\begin{equation}
\boldsymbol{z}_t^{T} = (1-t)\boldsymbol{z}_0^{T} + t\boldsymbol{z}_1^{T} 
                  = (1-t)\boldsymbol{x} + t\boldsymbol{\epsilon}.
\end{equation}
Let the velocity field for the T2V task be $\boldsymbol{v}_t^{T}(\boldsymbol{z}_t^{T},t)$, then $\boldsymbol{v}_t^{T}(\boldsymbol{z}_t^{T},t) = \frac{d\boldsymbol{z}_t^{T}}{dt} = \boldsymbol{\epsilon} - \boldsymbol{x}$. For the I2V task, let the source distribution sample be $\boldsymbol{z}_1^{I}$ and the target distribution sample be $\boldsymbol{z}_0^{I}$. Since FM imposes no constraints on the source and target distributions, we can modify the distributions to implicitly incorporate conditions and avoid conditional image leakage. $\boldsymbol{z}_1^{I}$ is modified to a linear mixture of the conditional image latents $\boldsymbol{s}=\mathcal{E}(\boldsymbol{S})$ and noise $\boldsymbol{\epsilon}$, and $\boldsymbol{z}_0^{I}$ is modified to a linear mixture of the conditional image latents $\boldsymbol{s}$ and the video $\boldsymbol{x}$, as follows:
\begin{align}
    \boldsymbol{z}_1^{I} &= \alpha \boldsymbol{s} + \beta \boldsymbol{\epsilon}, \\
    \boldsymbol{z}_0^{I} &= \gamma \boldsymbol{s} + \kappa \boldsymbol{x},
\end{align}
where $\alpha, \beta, \gamma, \kappa$ are undetermined constant numbers. We can compute the intermediate state $\boldsymbol{z}_t^{I}$ as:
\begin{equation}
    \boldsymbol{z}_t^{I} = (1-t)\boldsymbol{z}_0^{I} + t\boldsymbol{z}_1^{I} = \kappa \boldsymbol{z}_t^{T} + [\gamma + (\alpha - \gamma)t]\boldsymbol{s} + (\beta - \kappa)t\boldsymbol{\epsilon}.
\end{equation}
For the I2V task, let the velocity field be $\boldsymbol{v}_t^{I}(\boldsymbol{z}_t^{I},t)$, which can be expressed as:
\begin{equation}
    \boldsymbol{v}_t^{I}(\boldsymbol{z}_t^{I},t) = \frac{d\boldsymbol{z}_t^{I}}{dt} 
     = \kappa \boldsymbol{v}_t^{T}(\boldsymbol{z}_t^{T},t) + (\alpha - \gamma)\boldsymbol{s} + (\beta - \kappa)\boldsymbol{\epsilon}.
\end{equation}
When training an I2V model, we typically inherit the weight of the corresponding T2V model. A good initialization can leverage knowledge from the pre-trained weights as much as possible. It is observed that when $\alpha = \gamma$ and $\beta = \kappa = 1$, we have $\boldsymbol{v}_t^{I}(\boldsymbol{z}_t^{I},t) = \boldsymbol{v}_t^{T}(\boldsymbol{z}_t^{T},t)$, meaning the optimization objectives for I2V and T2V are structurally the same. In this case, $\boldsymbol{z}_t^{I}$ can be expressed as:
\begin{equation}
    \boldsymbol{z}_t^{I} = \boldsymbol{z}_t^{T} + \gamma \boldsymbol{s}.
\end{equation}
When $\gamma = 0$, we have $\boldsymbol{z}_t^{I}=\boldsymbol{z}_t^{T}$, meaning that without the conditional image as a condition, the model is equivalent to a T2V model. When $\gamma \ne 0$, the input of the denoiser incorporates conditional image information. In this case, we have $\boldsymbol{z}_1^{I} = \boldsymbol{\epsilon} - (-\gamma \boldsymbol{s})$ and $\boldsymbol{z}_0^{I} = \boldsymbol{x} - (-\gamma \boldsymbol{s})$, where $-\gamma \boldsymbol{s}$ can be viewed as the shifting of the latents in I2V task relative to T2V task and we aim to learn the conditional image information from the shifted latents.

Furthermore, $-\gamma$ can be viewed as a constant weight for $\boldsymbol{s}$. Since the effective information of each position varies, we can replace $-\gamma \boldsymbol{s}$ with $\boldsymbol{\phi}(\boldsymbol{s})$, where $\boldsymbol{\phi}(\cdot)$ is a learnable projection. Additionally, we can zero-initialize $\boldsymbol{\phi}(\cdot)$, ensuring that the input distribution of the denoiser is not disrupted at the start of training. Since $\boldsymbol{\phi}(\cdot)$ is a network that is independent of time $t$, we still have $\boldsymbol{v}_t^{I}(\boldsymbol{z}_t^{I},t) = \boldsymbol{v}_t^{T}(\boldsymbol{z}_t^{T},t)$. Now we obtain the ultimate form for the I2V method based on the latent shifting:
\begin{align}
    \boldsymbol{z}_1^I &= \boldsymbol{\epsilon} - \boldsymbol{\phi}(\boldsymbol{s}), \\
    \boldsymbol{z}_0^I &=  \boldsymbol{x} - \boldsymbol{\phi}(\boldsymbol{s}), \\
    \boldsymbol{z}_t^I &= \boldsymbol{z}_t^T - \boldsymbol{\phi}(\boldsymbol{s})
                                = (1-t)\boldsymbol{x} + t\boldsymbol{\epsilon} - \boldsymbol{\phi}(\boldsymbol{s}), \\
    \boldsymbol{v}_t^I(\boldsymbol{z}_t^I,t) &= \boldsymbol{v}_t^T(\boldsymbol{z}_t^T,t)
                                = \boldsymbol{\epsilon} - \boldsymbol{x}.
\end{align}

\subsection{Fourier Guidance}
\label{method:fourier_guidance}
During sampling, the latent shifting method needs to recover the information of $\boldsymbol{s}$ from the mixture of $\boldsymbol{\epsilon}$ and $\boldsymbol{\phi}(\boldsymbol{s})$. While recovery of low-frequency information like global color and shape is easier, high-frequency details like edges and contours are more challenging to recover accurately from the shifted noise. Therefore, models trained with the latent shifting require more time and data than the existing I2V paradigms to ensure the fidelity for the first frame.

We consider injecting the high-frequency information of $\boldsymbol{s}$ as additional input, aiming to address the challenge of learning high-frequency information. Since VAEs in latent diffusion models function similarly to AutoEncoders (AEs), we find that low-frequency and high-frequency information from the Fourier Transform in latent space resembles that in pixel space, but with significantly lower computational cost, as shown in the App.~\ref{app:fourier_guidance}. Since directly using high-frequency features leads to shortcutting by the model, we only retain the magnitudes of these features. Let $\boldsymbol{f}_{\text{high}}$ be the high-frequency magnitude filter of the Fourier Transform, we have:
\begin{equation}
    \boldsymbol{s}_{\text{high}} = \boldsymbol{f}_{\text{high}}(\boldsymbol{s}),
\end{equation}
where $\boldsymbol{s}_{\text{high}}$ means high-frequency magnitude features of $\boldsymbol{s}$. The detailed implementation of $\boldsymbol{f}_{\text{high}}$ can be found in the App.~\ref{app:fourier_guidance}. Then, $\boldsymbol{s}_{\text{high}}$ is concatenated with $\boldsymbol{z}_t^{I}$ along the channel dimension. After forwarding the embedding layers, we obtain the hidden states of the denoiser $\boldsymbol{H}$:
\begin{equation}
    \boldsymbol{H} = 
    \left[ \begin{matrix}
    	\boldsymbol{W}^{I}&		\boldsymbol{W}^{F} \\
    \end{matrix} \right] \left[ \begin{array}{c}
    	\boldsymbol{z}_t^{I} \\
    	\boldsymbol{s}_{\text{high}} \\
    \end{array} \right],
\end{equation}
where $[\cdot]$ denotes concatenation along channel dimension. Here, $\boldsymbol{W}^{I}$ represents patch embedding of the denoiser, and $\boldsymbol{W}^{F}$ is the embedding layer corresponding to $\boldsymbol{s}_{\text{high}}$. $\boldsymbol{W}^{F}$ is zero-initialized to ensure that the distribution of the hidden states remains unchanged at the beginning of training.

In summary, we can derive the loss function implemented by \method as follows:
\begin{equation}
    % \footnotesize
    \resizebox{0.92\textwidth}{!}{
        $\mathcal{L}_{\mathrm{Flash}}(\theta)=\mathbb{E}_{t\sim\mathcal{U}[0,1],\boldsymbol{\epsilon}\sim\mathcal{N}(\boldsymbol{0},\boldsymbol{I}),\boldsymbol{X}\sim q(\boldsymbol{X}),\boldsymbol{x}=\mathcal{E}(\boldsymbol{X}),\boldsymbol{s}=\mathcal{E}(\boldsymbol{X}[:,0])}\left[\left\|\boldsymbol{v}_\theta^{I}((1-t)\boldsymbol{x}+t\boldsymbol{\epsilon}-\boldsymbol{\phi}(\boldsymbol{s}),t,\boldsymbol{y},\boldsymbol{s}_{\mathrm{high}})-(\boldsymbol{\epsilon}-\boldsymbol{x})\right\|_2^2\right],$
    }
\end{equation}
where $\boldsymbol{y}$ is the text embedding, and $\boldsymbol{v}_\theta^{I}$ is the denoiser excluding $\boldsymbol{\phi}$.
\section{Experiment}
In this section, we first introduce the experimental setup in Sec.~\ref{exp:experimental_setup}. Then, in Sec.~\ref{exp:comparisons}, we compare \method with other I2V methods from both quantitative and qualitative perspectives. In addition, in Sec.~\ref{exp:ablation_study}, we present the results of the ablation experiments to demonstrate the effectiveness of \method. Finally, in Sec.~\ref{exp:analysis}, we analyze the functions of the modules in \method.
\subsection{Experimental Setup}
\label{exp:experimental_setup}
\begin{table}[t!]
    \samepage
    \centering
    \caption{{\bf Vbench-I2V results}. We compare the performance of various methods on Vbench-I2V. It can be observed that, despite having the fewest parameters, \method achieves comparable scores to models with larger parameter sizes. The dynamic degree score of \method significantly surpasses that of other methods. All scores are presented as percentages (\%). $\dagger$ indicates testing using recaptioning image-text pairs on Vbench-I2V. For further details, see the App.~\ref{app:recaptioning_vbench_i2v}.
    }
    \resizebox{\textwidth}{!}{%
    \setlength\tabcolsep{3pt}
    \begin{tabular}{lccccccccccccc}
        \toprule
        Model & I2V Paradigm & \makecell{Subject\\ Consistency$\uparrow$} & \makecell{Background\\ Consistency$\uparrow$} & \makecell{Motion\\ Smoothness$\uparrow$} & \makecell{Dynamic\\ Degree$\uparrow$} & \makecell{Aesthetic\\ Quality$\uparrow$} & \makecell{Imaging\\ Quality$\uparrow$} & \makecell{I2V Subject\\ Consistency$\uparrow$} & \makecell{I2V Background\\ Consistency$\uparrow$} \\
        \midrule
        SVD-XT-1.0 (1.5B) & Repeating Concat and Adding Noise & 95.52 & 96.61 & 98.09 & 52.36 & 60.15 & 69.80 & 97.52 & 97.63 \\
        SVD-XT-1.1 (1.5B) & Repeating Concat and Adding Noise & 95.42 & 96.77 & 98.12 & 43.17 & 60.23 & 70.23 & 97.51 & 97.62 \\
        SEINE-512x512 (1.8B) & Inpainting & 95.28 & 97.12 & 97.12 & 27.07 & 64.55 & {\bf 71.39} & 97.15 & 96.94 \\
        CogVideoX-5B-I2V & Zero-padding Concat and Adding Noise & 94.34 & 96.42 & 98.40 & 33.17 & 61.87 & 70.01 & 97.19 & 96.74 \\
        Wan2.1-I2V-14B-720P & Inpainting & 94.86 & 97.07 & 97.90 & 51.38 & {\bf 64.75} & 70.44 & 96.95 & 96.44 \\
        \midrule
        CogVideoX1.5-5B-I2V$^\dagger$ & Zero-padding Concat and Adding Noise & 95.04 & 96.52 & {\bf 98.47} & 37.48 & {\bf 62.68} & {\bf 70.99} & 97.78 & 98.73 \\
        Wan2.1-I2V-14B-480P$^\dagger$ & Inpainting & {\bf 95.68} & {\bf 97.44} & 98.46 & 45.20 & 61.44 & 70.37 & {\bf 97.83} & {\bf 99.08} \\
        \rowcolor[HTML]{dae7ed} {\bf FlashI2V$^\dagger$ (1.3B)} & {\bf FlashI2V} & 95.13 & 96.36 & 98.35 & {\bf 53.01} & 62.34 & 69.41 & 97.67 & 98.72 \\
        \bottomrule
    \end{tabular}
    }
    \label{tab:experiment_comparisons}
    \vspace{-0.5em}
\end{table}
\begin{figure}[t!]
    \samepage
    \centering
    \includegraphics[width=\linewidth]{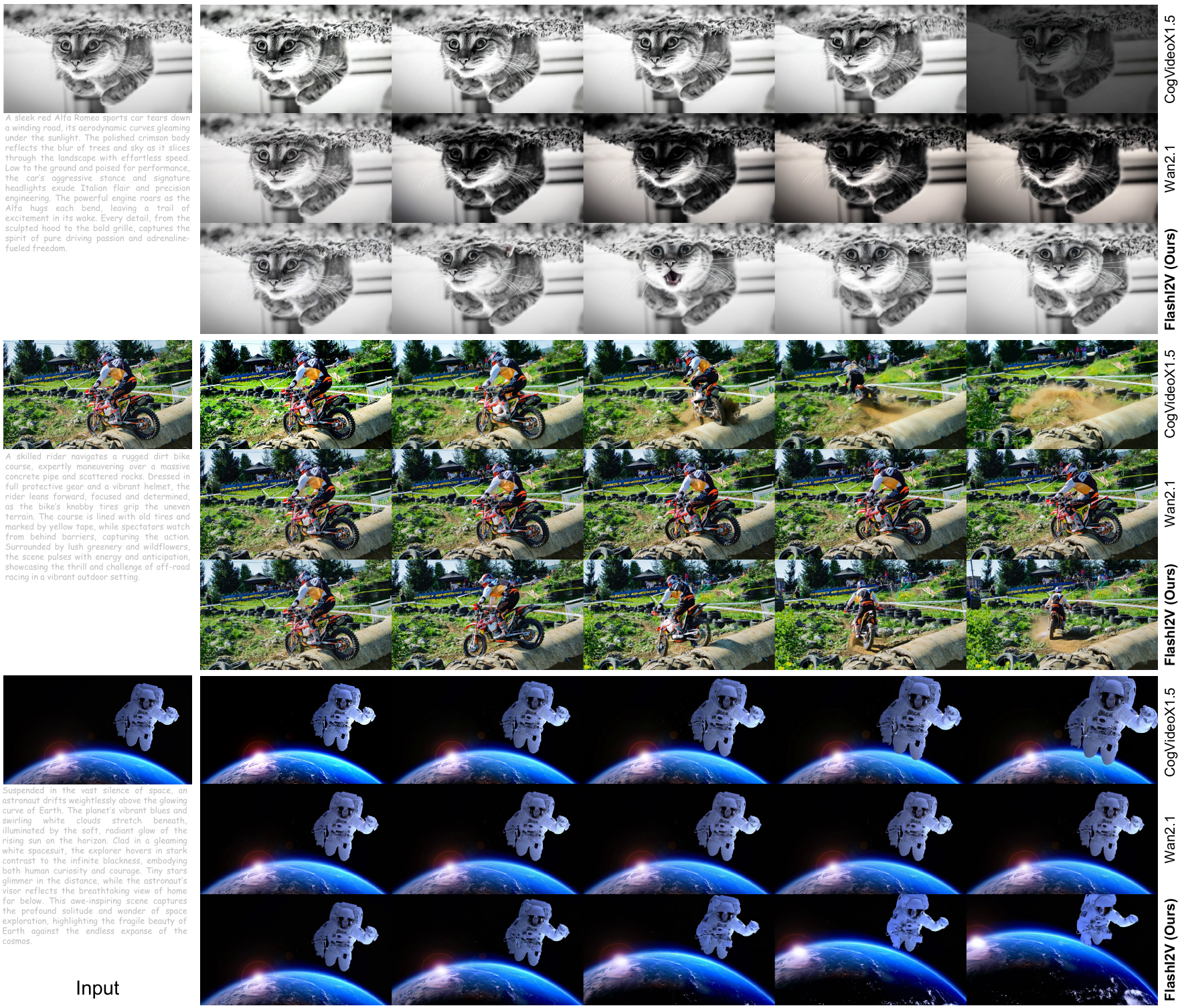}
    \caption{{\bf Method Comparison}. We compare the quantitative performance of \method (1.3B) with CogVideoX1.5-5B-I2V~\citep{cogvideox} and Wan2.1-I2V-14B-480P~\citep{wan}. We observe that CogVideoX1.5 and Wan2.1 exhibit color inconsistency. Additionally, Wan2.1 tends to produce extremely slow-motion or even static videos. Thanks to the avoidance of conditional image leakage, \method effectively resolves these performance degradation issues.}
    \label{fig:experiment_comparison}
    \vspace{-2em}
\end{figure}
\textbf{Training Setup}. In the comparisons, we train a model for 84K steps on 20M high-quality video data collected internally, following the collection and processing pipeline described in Open-Sora Plan~\citep{open_sora_plan}. For each video, we randomly sample 49 frames at a fixed fps of 16, with a resolution of $\text{480} \times \text{832}$. We initialize the model from the Wan2.1-T2V-1.3B~\citep{wan} model. The learnable projection is implemented using two layers of Conv3D~\citep{conv3d} and SiLU~\citep{silu}, and the Fourier embedding layer is implemented in the same way as the patch embedding, both with zero initialization. During training, the first frame of each video serves as the conditional image. The cutoff frequency percentile of the Fourier Transform is sampled from $\mathcal{U}[0.05, 0.95]$. The text prompt is dropped with a probability of 0.1. We use a batch size of 256, a learning rate of 4e-5, a weight decay of 1e-2, and the AdamW optimizer with $\beta_1$ set to 0.9, $\beta_2$ set to 0.999, and $\epsilon$ set to 1e-15. The weights are updated using Exponential Moving Average (EMA) with a decay of 0.9999. In the ablation study, all models involved in the comparison are initialized from Wan2.1-T2V-1.3B. We select a 2M subset as the training set, use a learning rate of 2e-5, a batch size of 64, and 30K training steps, while keeping the other settings unchanged.
\textbf{Sampling Setup}. For sampling, we use the Discrete Euler Sampler with a sigma shifting strategy as in HunyuanVideo~\citep{hunyuanvideo}, a shifting coefficient of 7.0, classifier-free guidance set to 5.0, 50 sampling steps, and a cutoff frequency percentile set to 0.1.

\textbf{Evaluation}. In the comparisons, we use a fixed 49 frames and the default resolution, and we utilize all Vbench-I2V~\citep{vbench,vbench2} metrics except for Camera Motion as evaluation metrics. In addition, we use ChatGPT~\citep{gpt4} to rewrite the short prompts of Vbench-I2V in order to obtain more accurate evaluation results. For further details, please refer to the App.~\ref{app:recaptioning_vbench_i2v}. In the ablation study, we randomly select 1,000 videos from the HD subset of OpenVid-1M~\citep{openvid} as the validation set, calculating the chunk-wise FVD for each setting.
\subsection{Main Results}
\label{exp:comparisons}
\textbf{Quantitative results.} We compare the performance of different methods on Vbench-I2V, as shown in Tab.~\ref{tab:experiment_comparisons}. Because of preventing conditional image leakage, \method achieves a significantly higher dynamic degree score across all methods. In other metrics, \method is quite close to CogVideoX1.5-5B-I2V and Wan2.1-I2V-14B-480P with larger parameter sizes. It outperforms CogVideoX1.5 in the Subject Consistency metric and exceeds Wan2.1 in the Aesthetic Quality metric.

\textbf{Qualitative results.} As shown in Fig.~\ref{fig:experiment_comparison}, we compare the qualitative performance of different methods. Due to the impact of conditional image leakage, CogVideoX1.5 and Wan2.1 exhibit issues such as color inconsistency and slow motion in some samples. Wan2.1 even produces completely static videos. In contrast, \method generates videos with larger motion and adheres more closely to physical laws.

\subsection{Ablation Study}
\begin{figure}[t!]
    \centering
    \begin{subfigure}[h]{\textwidth}
        \centering
        \includegraphics[width=0.96\textwidth]{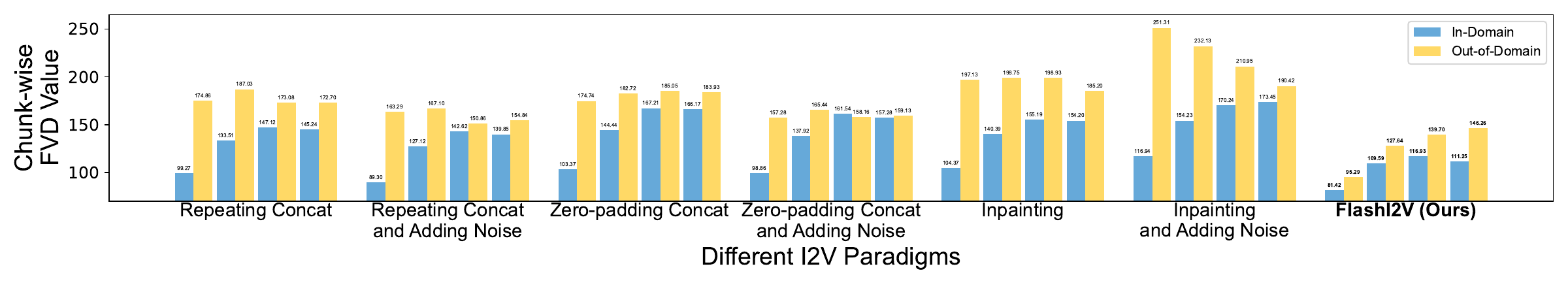}
        \caption{Chunk-wise FVD Variation Patterns Across Various I2V Paradigms}
        \label{fig:experiment_ablation_study_overfitting}
    \end{subfigure}
    \begin{subfigure}[h]{0.32\textwidth}
        \centering
        \includegraphics[width=\textwidth]{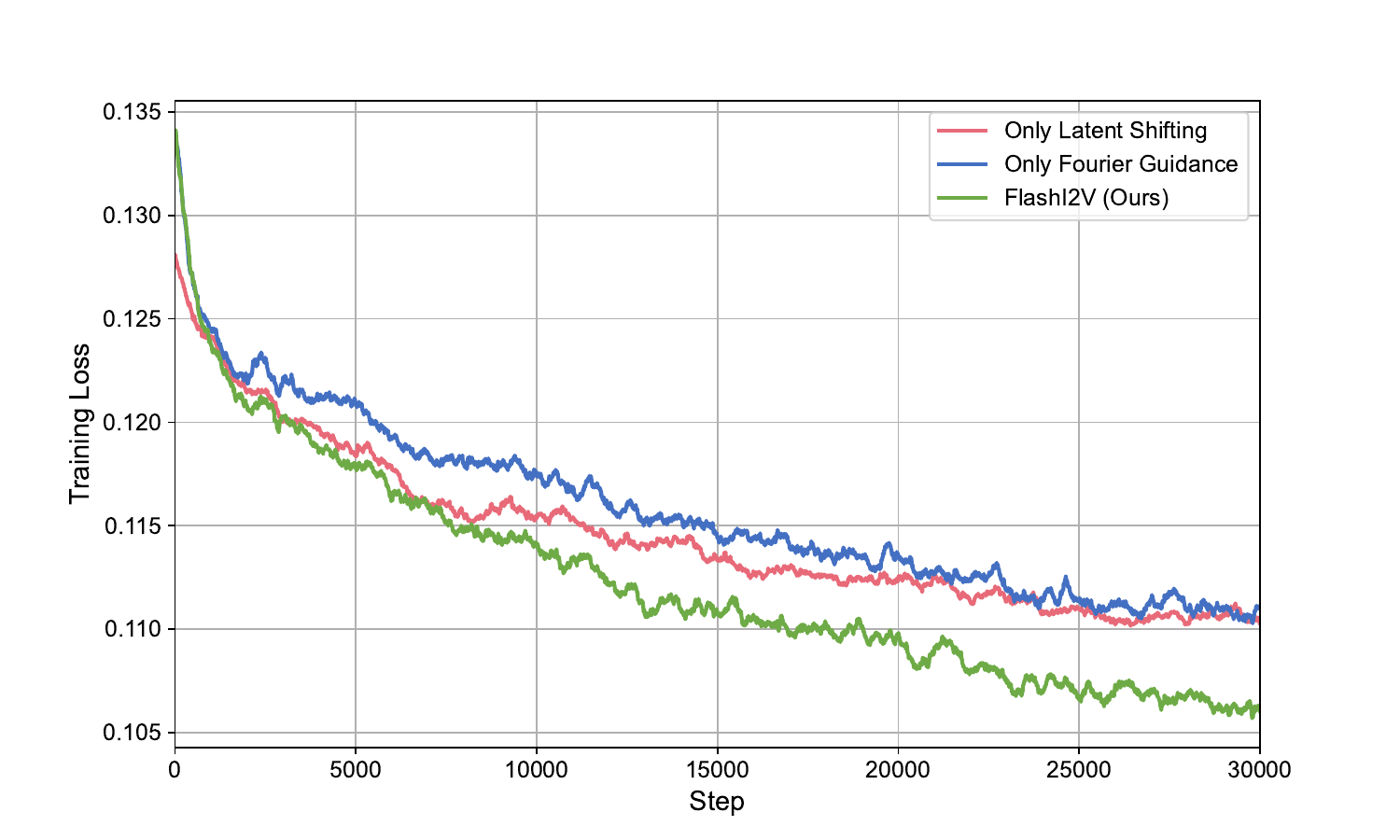}
        \caption{Training Loss}
        \label{fig:experiments_ablation_study_training_loss}
    \end{subfigure}
    \begin{subfigure}[h]{0.66\textwidth}
        \centering
        \includegraphics[width=\textwidth]{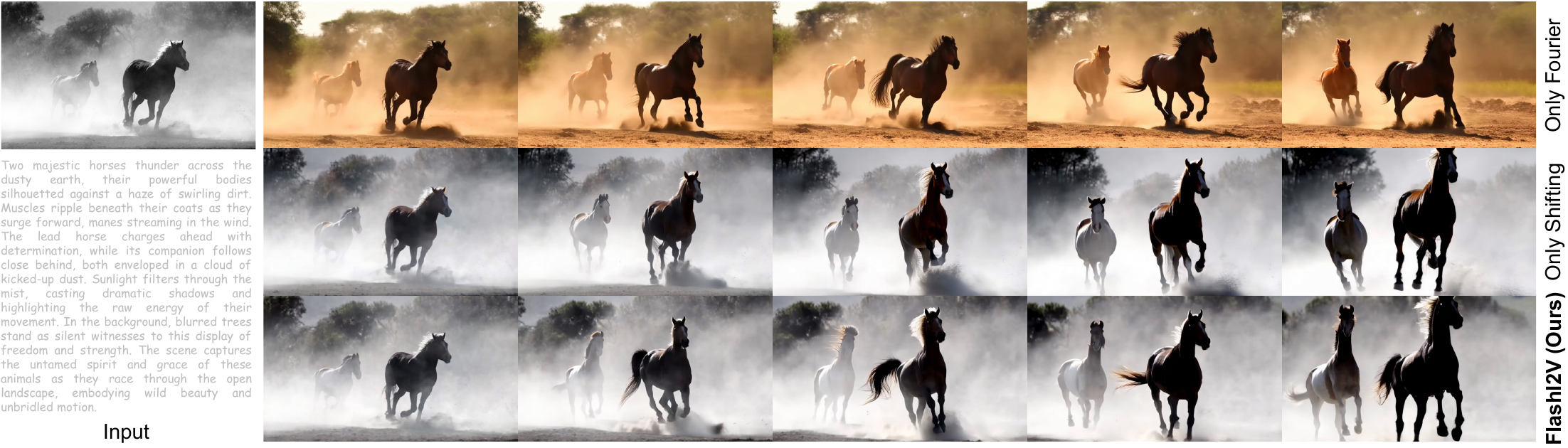}
        \caption{Qualitative Results}
        \label{fig:experiments_ablation_study_samples}
    \end{subfigure}
    \caption{{\bf Ablation Study}. (a) Comparing the chunk-wise FVD variation patterns of different I2V paradigms on both the training and validation sets, it is observed that only \method exhibits the same time-increasing FVD variation pattern in both sets. This suggests that only \method is capable of applying the generation law learned from in-domain data to out-of-domain data. Additionally, \method has the lowest out-of-domain FVD, demonstrating its performance advantage. (b) From the training loss, we can observe that Fourier guidance accelerates the convergence of latent shifting. (c) Fourier guidance alone causes color deviation, while latent shifting alone leads to mismatched details. \method achieves consistency in both color and details.}
    \label{fig:experiment_ablation_study}
    \vspace{-1em}
\end{figure}
\begin{figure}[t]
    \centering
    \begin{subfigure}[h]{\textwidth}
        \centering
        \includegraphics[width=\textwidth]{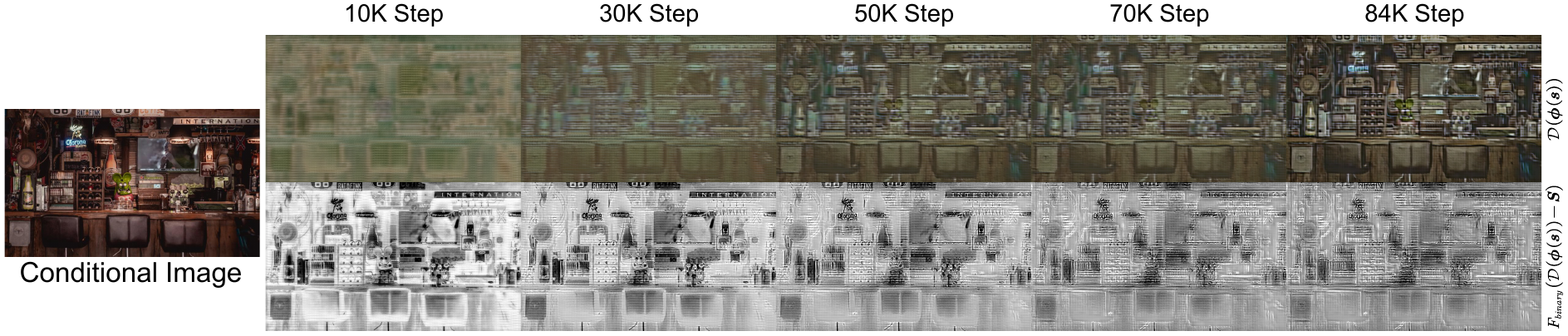}
        \caption{Encoded Features of Latent Shifting}
        \label{fig:experiment_analysis_latent_shifting}
    \end{subfigure}
    \begin{subfigure}[h]{\textwidth}
        \centering
        \includegraphics[width=\textwidth]{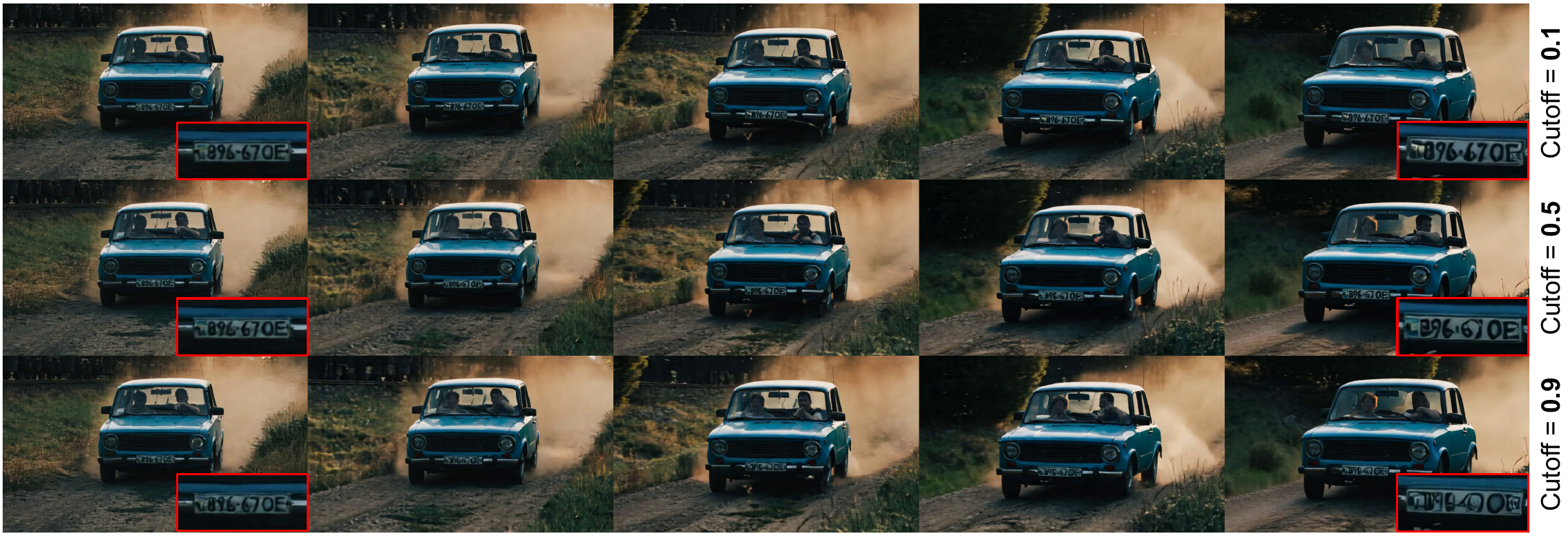}
        \caption{The Influence of Fourier Guidance on Generation Details}
        \label{fig:experiment_analysis_fourier_guidance}
    \end{subfigure}
    \caption{{\bf Analysis of latent shifting and fourier guidance}. (a) As training progresses, $\boldsymbol{\phi}(\cdot)$ gradually emphasizes the detailed information in the conditional image. (b) When a lower cutoff frequency percentile is used, more high-frequency information is injected. When the cutoff frequency percentile is set to 0.1, the graphical text at the end of the video remains unchanged, while with the cutoff frequency percentile set to 0.9, the graphical text becomes unrecognizable.}
    \vspace{-1.5em}
\end{figure}
\label{exp:ablation_study}
{\bf Generalization to out-of-domain data}. We compare the chunk-wise FVD variation patterns of different I2V paradigms on in-domain and out-of-domain data, as shown in Fig.~\ref{fig:experiment_ablation_study_overfitting}. The pseudocode implementations of the different paradigms can be found in the App.~\ref{app:code_implementation}. Each generated video is divided into four temporal chunks over time with an equal interval. Apart from \method, other paradigms concatenate the full information of the conditional image with the noisy latents. The paradigms named with "Adding Noise" add a small amount of noise to the conditional image latents, similar to the implementation in CogVideoX~\citep{cogvideox}. We observe that only \method exhibits the same chunk-wise FVD variation pattern on both in-domain and out-of-domain data, indicating that the generation law learned by \method on in-domain data generalizes well to out-of-domain data. In contrast, other paradigms show inconsistent chunk-wise FVD variation patterns between in-domain and out-of-domain data, suggesting the leakage caused by shortcutting conditional images. Moreover, our method achieves the lowest FVD on out-of-domain data, meaning it has the best performance across different I2V paradigms.

{\bf The functions of various modules in \method}. To investigate the effectiveness of latent shifting and Fourier guidance, we conduct detailed ablation experiments. From a quantitative perspective, as shown in Fig.~\ref{fig:experiments_ablation_study_training_loss}, \method achieves a faster decline in training loss by incorporating Fourier guidance compared to using latent shifting alone, indicating that Fourier guidance effectively accelerates the convergence of latent shifting. From a qualitative perspective, we compare the performance after removing different modules in Fig.~\ref{fig:experiments_ablation_study_samples}. When using only Fourier guidance, the generated video maintains high-frequency content consistent with the conditional image. However, due to receiving only the magnitude features, it fails to produce correct colors with only Fourier guidance. With only latent shifting, the generated scene aligns with the conditional image but lacks satisfactory fidelity in local details. \method successfully achieves both global and local fidelity.

\vspace{-0.5em}
\subsection{Analysis}
\label{exp:analysis}

{\bf Features encoded through latent shifting}. Since $\boldsymbol{\phi}(\cdot)$ performs a shifting operation on latents, its encoded features are meaningful in the latent space. As shown in Fig.~\ref{fig:experiment_analysis_latent_shifting}, we visualize the features encoded by $\boldsymbol{\phi}(\cdot)$ in the pixel space by VAE decoding and compute the relative difference between $\mathcal{D}(\boldsymbol{\phi}(\boldsymbol{s}))$ and $\boldsymbol{S}$ in the pixel space, represented as a binary image. As training progresses, the features encoded by $\boldsymbol{\phi}(\cdot)$ become richer, and $\boldsymbol{\phi}(\boldsymbol{s})$ emphasizes high-frequency representations compared to $\boldsymbol{s}$, resulting in gradually improved fidelity of the model during training.

{\bf Adjustable generation details}. At different cutoff frequency percentiles, Fourier guidance can provide varying detail levels. As shown in Fig.~\ref{fig:experiment_analysis_fourier_guidance}, we compare the influence of changing cutoff frequency percentiles on the generated details. A lower cutoff frequency percentile means injecting richer high-frequency details, resulting in finer generation and better detail preservation during the entire video, especially for small-scale regions like graphical text. 

\vspace{-0.5em}
\section{Conclusion}
Existing I2V paradigms cannot avoid conditional image leakage, leading to performance degradation. We propose \method, which implicitly introduces conditions through latent shifting. Additionally, we utilize high-frequency magnitude features extracted by the Fourier Transform as guidance to accelerate the convergence. Experimental results show that \method demonstrates the best generalization and performance on out-of-domain data. With only 1.3B parameters, \method achieves the best dynamic degree score across various methods on Vbench-I2V.

\bibliography{iclr2026_conference}

@article{open_sora_plan,
  title={Open-sora plan: Open-source large video generation model},
  author={Lin, Bin and Ge, Yunyang and Cheng, Xinhua and Li, Zongjian and Zhu, Bin and Wang, Shaodong and He, Xianyi and Ye, Yang and Yuan, Shenghai and Chen, Liuhan and others},
  journal={arXiv preprint arXiv:2412.00131},
  year={2024}
}

@article{open_sora,
  title={Open-sora: Democratizing efficient video production for all},
  author={Zheng, Zangwei and Peng, Xiangyu and Yang, Tianji and Shen, Chenhui and Li, Shenggui and Liu, Hongxin and Zhou, Yukun and Li, Tianyi and You, Yang},
  journal={arXiv preprint arXiv:2412.20404},
  year={2024}
}

@article{magictime,
  title={Magictime: Time-lapse video generation models as metamorphic simulators},
  author={Yuan, Shenghai and Huang, Jinfa and Shi, Yujun and Xu, Yongqi and Zhu, Ruijie and Lin, Bin and Cheng, Xinhua and Yuan, Li and Luo, Jiebo},
  journal={IEEE Transactions on Pattern Analysis and Machine Intelligence},
  year={2025},
  publisher={IEEE}
}

@article{chronomagic,
  title={Chronomagic-bench: A benchmark for metamorphic evaluation of text-to-time-lapse video generation},
  author={Yuan, Shenghai and Huang, Jinfa and Xu, Yongqi and Liu, Yaoyang and Zhang, Shaofeng and Shi, Yujun and Zhu, Rui-Jie and Cheng, Xinhua and Luo, Jiebo and Yuan, Li},
  journal={Advances in Neural Information Processing Systems},
  volume={37},
  pages={21236--21270},
  year={2024}
}

@article{opens2v,
  title={Opens2v-nexus: A detailed benchmark and million-scale dataset for subject-to-video generation},
  author={Yuan, Shenghai and He, Xianyi and Deng, Yufan and Ye, Yang and Huang, Jinfa and Lin, Bin and Luo, Jiebo and Yuan, Li},
  journal={arXiv preprint arXiv:2505.20292},
  year={2025}
}

@article{videocrafter1,
  title={Videocrafter1: Open diffusion models for high-quality video generation},
  author={Chen, Haoxin and Xia, Menghan and He, Yingqing and Zhang, Yong and Cun, Xiaodong and Yang, Shaoshu and Xing, Jinbo and Liu, Yaofang and Chen, Qifeng and Wang, Xintao and others},
  journal={arXiv preprint arXiv:2310.19512},
  year={2023}
}

@inproceedings{videocrafter2,
  title={Videocrafter2: Overcoming data limitations for high-quality video diffusion models},
  author={Chen, Haoxin and Zhang, Yong and Cun, Xiaodong and Xia, Menghan and Wang, Xintao and Weng, Chao and Shan, Ying},
  booktitle={Proceedings of the IEEE/CVF Conference on Computer Vision and Pattern Recognition},
  pages={7310--7320},
  year={2024}
}

@article{lavie,
  title={Lavie: High-quality video generation with cascaded latent diffusion models},
  author={Wang, Yaohui and Chen, Xinyuan and Ma, Xin and Zhou, Shangchen and Huang, Ziqi and Wang, Yi and Yang, Ceyuan and He, Yinan and Yu, Jiashuo and Yang, Peiqing and others},
  journal={International Journal of Computer Vision},
  volume={133},
  number={5},
  pages={3059--3078},
  year={2025},
  publisher={Springer}
}

@article{animatediff,
  title={Animatediff: Animate your personalized text-to-image diffusion models without specific tuning},
  author={Guo, Yuwei and Yang, Ceyuan and Rao, Anyi and Liang, Zhengyang and Wang, Yaohui and Qiao, Yu and Agrawala, Maneesh and Lin, Dahua and Dai, Bo},
  journal={arXiv preprint arXiv:2307.04725},
  year={2023}
}

@article{latte,
  title={Latte: Latent diffusion transformer for video generation},
  author={Ma, Xin and Wang, Yaohui and Jia, Gengyun and Chen, Xinyuan and Liu, Ziwei and Li, Yuan-Fang and Chen, Cunjian and Qiao, Yu},
  journal={arXiv preprint arXiv:2401.03048},
  year={2024}
}

@inproceedings{dit,
  title={Scalable diffusion models with transformers},
  author={Peebles, William and Xie, Saining},
  booktitle={Proceedings of the IEEE/CVF international conference on computer vision},
  pages={4195--4205},
  year={2023}
}

@inproceedings{lightingdit,
  title={Reconstruction vs. generation: Taming optimization dilemma in latent diffusion models},
  author={Yao, Jingfeng and Yang, Bin and Wang, Xinggang},
  booktitle={Proceedings of the Computer Vision and Pattern Recognition Conference},
  pages={15703--15712},
  year={2025}
}

@inproceedings{unet,
  title={U-net: Convolutional networks for biomedical image segmentation},
  author={Ronneberger, Olaf and Fischer, Philipp and Brox, Thomas},
  booktitle={International Conference on Medical image computing and computer-assisted intervention},
  pages={234--241},
  year={2015},
  organization={Springer}
}

@article{ddpm,
  title={Denoising diffusion probabilistic models},
  author={Ho, Jonathan and Jain, Ajay and Abbeel, Pieter},
  journal={Advances in neural information processing systems},
  volume={33},
  pages={6840--6851},
  year={2020}
}

@article{ddim,
  title={Denoising diffusion implicit models},
  author={Song, Jiaming and Meng, Chenlin and Ermon, Stefano},
  journal={arXiv preprint arXiv:2010.02502},
  year={2020}
}

@article{score_diffusion,
  title={Score-based generative modeling through stochastic differential equations},
  author={Song, Yang and Sohl-Dickstein, Jascha and Kingma, Diederik P and Kumar, Abhishek and Ermon, Stefano and Poole, Ben},
  journal={arXiv preprint arXiv:2011.13456},
  year={2020}
}

@article{flow_matching,
  title={Flow matching for generative modeling},
  author={Lipman, Yaron and Chen, Ricky TQ and Ben-Hamu, Heli and Nickel, Maximilian and Le, Matt},
  journal={arXiv preprint arXiv:2210.02747},
  year={2022}
}

@inproceedings{animate_anyone,
  title={Animate anyone: Consistent and controllable image-to-video synthesis for character animation},
  author={Hu, Li},
  booktitle={Proceedings of the IEEE/CVF Conference on Computer Vision and Pattern Recognition},
  pages={8153--8163},
  year={2024}
}

@inproceedings{consisid,
  title={Identity-preserving text-to-video generation by frequency decomposition},
  author={Yuan, Shenghai and Huang, Jinfa and He, Xianyi and Ge, Yunyang and Shi, Yujun and Chen, Liuhan and Luo, Jiebo and Yuan, Li},
  booktitle={Proceedings of the Computer Vision and Pattern Recognition Conference},
  pages={12978--12988},
  year={2025}
}

@article{anyi2v,
  title={AnyI2V: Animating Any Conditional Image with Motion Control},
  author={Li, Ziye and Luo, Hao and Shuai, Xincheng and Ding, Henghui},
  journal={arXiv preprint arXiv:2507.02857},
  year={2025}
}

@article{rectified_flow,
  title={Rectified flow: A marginal preserving approach to optimal transport},
  author={Liu, Qiang},
  journal={arXiv preprint arXiv:2209.14577},
  year={2022}
}

@article{viewcrafter,
  title={Viewcrafter: Taming video diffusion models for high-fidelity novel view synthesis},
  author={Yu, Wangbo and Xing, Jinbo and Yuan, Li and Hu, Wenbo and Li, Xiaoyu and Huang, Zhipeng and Gao, Xiangjun and Wong, Tien-Tsin and Shan, Ying and Tian, Yonghong},
  journal={arXiv preprint arXiv:2409.02048},
  year={2024}
}

@inproceedings{query_transformer,
  title={Blip-2: Bootstrapping language-image pre-training with frozen image encoders and large language models},
  author={Li, Junnan and Li, Dongxu and Savarese, Silvio and Hoi, Steven},
  booktitle={International conference on machine learning},
  pages={19730--19742},
  year={2023},
  organization={PMLR}
}

@inproceedings{conv3d,
  title={Learning spatiotemporal features with 3d convolutional networks},
  author={Tran, Du and Bourdev, Lubomir and Fergus, Rob and Torresani, Lorenzo and Paluri, Manohar},
  booktitle={Proceedings of the IEEE international conference on computer vision},
  pages={4489--4497},
  year={2015}
}

@article{silu,
  title={Sigmoid-weighted linear units for neural network function approximation in reinforcement learning},
  author={Elfwing, Stefan and Uchibe, Eiji and Doya, Kenji},
  journal={Neural networks},
  volume={107},
  pages={3--11},
  year={2018},
  publisher={Elsevier}
}

@article{vae,
  title={Auto-encoding variational bayes},
  author={Kingma, Diederik P and Welling, Max},
  journal={arXiv preprint arXiv:1312.6114},
  year={2013}
}

@inproceedings{clip,
  title={Learning transferable visual models from natural language supervision},
  author={Radford, Alec and Kim, Jong Wook and Hallacy, Chris and Ramesh, Aditya and Goh, Gabriel and Agarwal, Sandhini and Sastry, Girish and Askell, Amanda and Mishkin, Pamela and Clark, Jack and others},
  booktitle={International conference on machine learning},
  pages={8748--8763},
  year={2021},
  organization={PmLR}
}

@article{sora,
  title={Video generation models as world simulators},
  author={Tim Brooks and Bill Peebles and Connor Holmes and Will DePue and Yufei Guo and Li Jing and David Schnurr and Joe Taylor and Troy Luhman and Eric Luhman and Clarence Ng and Ricky Wang and Aditya Ramesh},
  year={2024},
  url={https://openai.com/research/video-generation-models-as-world-simulators},
}

@article{gpt4,
  title={Gpt-4 technical report},
  author={Achiam, Josh and Adler, Steven and Agarwal, Sandhini and Ahmad, Lama and Akkaya, Ilge and Aleman, Florencia Leoni and Almeida, Diogo and Altenschmidt, Janko and Altman, Sam and Anadkat, Shyamal and others},
  journal={arXiv preprint arXiv:2303.08774},
  year={2023}
}

@article{openvid,
  title={Openvid-1m: A large-scale high-quality dataset for text-to-video generation},
  author={Nan, Kepan and Xie, Rui and Zhou, Penghao and Fan, Tiehan and Yang, Zhenheng and Chen, Zhijie and Li, Xiang and Yang, Jian and Tai, Ying},
  journal={arXiv preprint arXiv:2407.02371},
  year={2024}
}

@inproceedings{vbench,
  title={Vbench: Comprehensive benchmark suite for video generative models},
  author={Huang, Ziqi and He, Yinan and Yu, Jiashuo and Zhang, Fan and Si, Chenyang and Jiang, Yuming and Zhang, Yuanhan and Wu, Tianxing and Jin, Qingyang and Chanpaisit, Nattapol and others},
  booktitle={Proceedings of the IEEE/CVF Conference on Computer Vision and Pattern Recognition},
  pages={21807--21818},
  year={2024}
}

@article{vbench2,
  title={Vbench-2.0: Advancing video generation benchmark suite for intrinsic faithfulness},
  author={Zheng, Dian and Huang, Ziqi and Liu, Hongbo and Zou, Kai and He, Yinan and Zhang, Fan and Zhang, Yuanhan and He, Jingwen and Zheng, Wei-Shi and Qiao, Yu and others},
  journal={arXiv preprint arXiv:2503.21755},
  year={2025}
}

@article{wan,
  title={Wan: Open and advanced large-scale video generative models},
  author={Wan, Team and Wang, Ang and Ai, Baole and Wen, Bin and Mao, Chaojie and Xie, Chen-Wei and Chen, Di and Yu, Feiwu and Zhao, Haiming and Yang, Jianxiao and others},
  journal={arXiv preprint arXiv:2503.20314},
  year={2025}
}

@article{hunyuanvideo,
  title={Hunyuanvideo: A systematic framework for large video generative models},
  author={Kong, Weijie and Tian, Qi and Zhang, Zijian and Min, Rox and Dai, Zuozhuo and Zhou, Jin and Xiong, Jiangfeng and Li, Xin and Wu, Bo and Zhang, Jianwei and others},
  journal={arXiv preprint arXiv:2412.03603},
  year={2024}
}

@article{cogvideox,
  title={Cogvideox: Text-to-video diffusion models with an expert transformer},
  author={Yang, Zhuoyi and Teng, Jiayan and Zheng, Wendi and Ding, Ming and Huang, Shiyu and Xu, Jiazheng and Yang, Yuanming and Hong, Wenyi and Zhang, Xiaohan and Feng, Guanyu and others},
  journal={arXiv preprint arXiv:2408.06072},
  year={2024}
}

@article{stable_video_diffusion,
  title={Stable video diffusion: Scaling latent video diffusion models to large datasets},
  author={Blattmann, Andreas and Dockhorn, Tim and Kulal, Sumith and Mendelevitch, Daniel and Kilian, Maciej and Lorenz, Dominik and Levi, Yam and English, Zion and Voleti, Vikram and Letts, Adam and others},
  journal={arXiv preprint arXiv:2311.15127},
  year={2023}
}

@inproceedings{dynamicrafter,
  title={Dynamicrafter: Animating open-domain images with video diffusion priors},
  author={Xing, Jinbo and Xia, Menghan and Zhang, Yong and Chen, Haoxin and Yu, Wangbo and Liu, Hanyuan and Liu, Gongye and Wang, Xintao and Shan, Ying and Wong, Tien-Tsin},
  booktitle={European Conference on Computer Vision},
  pages={399--417},
  year={2024},
  organization={Springer}
}

@article{i2vgen_xl,
  title={I2vgen-xl: High-quality image-to-video synthesis via cascaded diffusion models},
  author={Zhang, Shiwei and Wang, Jiayu and Zhang, Yingya and Zhao, Kang and Yuan, Hangjie and Qin, Zhiwu and Wang, Xiang and Zhao, Deli and Zhou, Jingren},
  journal={arXiv preprint arXiv:2311.04145},
  year={2023}
}

@article{consisti2v,
  title={Consisti2v: Enhancing visual consistency for image-to-video generation},
  author={Ren, Weiming and Yang, Huan and Zhang, Ge and Wei, Cong and Du, Xinrun and Huang, Wenhao and Chen, Wenhu},
  journal={arXiv preprint arXiv:2402.04324},
  year={2024}
}

@inproceedings{seine,
  title={Seine: Short-to-long video diffusion model for generative transition and prediction},
  author={Chen, Xinyuan and Wang, Yaohui and Zhang, Lingjun and Zhuang, Shaobin and Ma, Xin and Yu, Jiashuo and Wang, Yali and Lin, Dahua and Qiao, Yu and Liu, Ziwei},
  booktitle={The Twelfth International Conference on Learning Representations},
  year={2023}
}

@article{easyanimate,
  title={Easyanimate: A high-performance long video generation method based on transformer architecture},
  author={Xu, Jiaqi and Zou, Xinyi and Huang, Kunzhe and Chen, Yunkuo and Liu, Bo and Cheng, MengLi and Shi, Xing and Huang, Jun},
  journal={arXiv preprint arXiv:2405.18991},
  year={2024}
}

@article{improving_flow_matching,
  title={Improving and generalizing flow-based generative models with minibatch optimal transport},
  author={Tong, Alexander and Fatras, Kilian and Malkin, Nikolay and Huguet, Guillaume and Zhang, Yanlei and Rector-Brooks, Jarrid and Wolf, Guy and Bengio, Yoshua},
  journal={arXiv preprint arXiv:2302.00482},
  year={2023}
}

@misc{CNFs,
	author={Chen, Ricky T. Q.},
	title={torchdiffeq},
	year={2018},
	url={https://github.com/rtqichen/torchdiffeq},
}

@article{fantasyid,
  title={Fantasyid: Face knowledge enhanced id-preserving video generation},
  author={Zhang, Yunpeng and Wang, Qiang and Jiang, Fan and Fan, Yaqi and Xu, Mu and Qi, Yonggang},
  journal={arXiv preprint arXiv:2502.13995},
  year={2025}
}

@misc{kling,
  author       = {Kling AI},
  year         = 2025,
  url          = {https://klingai.com},
}

@article{conditional_image_leakage,
  title={Identifying and solving conditional image leakage in image-to-video diffusion model},
  author={Zhao, Min and Zhu, Hongzhou and Xiang, Chendong and Zheng, Kaiwen and Li, Chongxuan and Zhu, Jun},
  journal={Advances in Neural Information Processing Systems},
  volume={37},
  pages={30300--30326},
  year={2024}
}

@article{ALG,
  title={Enhancing Motion Dynamics of Image-to-Video Models via Adaptive Low-Pass Guidance},
  author={Choi, June Suk and Lee, Kyungmin and Yu, Sihyun and Choi, Yisol and Shin, Jinwoo and Lee, Kimin},
  journal={arXiv preprint arXiv:2506.08456},
  year={2025}
}

@inproceedings{panda70m,
  title={Panda-70m: Captioning 70m videos with multiple cross-modality teachers},
  author={Chen, Tsai-Shien and Siarohin, Aliaksandr and Menapace, Willi and Deyneka, Ekaterina and Chao, Hsiang-wei and Jeon, Byung Eun and Fang, Yuwei and Lee, Hsin-Ying and Ren, Jian and Yang, Ming-Hsuan and others},
  booktitle={Proceedings of the IEEE/CVF Conference on Computer Vision and Pattern Recognition},
  pages={13320--13331},
  year={2024}
}

@article{languagebind,
  title={Languagebind: Extending video-language pretraining to n-modality by language-based semantic alignment},
  author={Zhu, Bin and Lin, Bin and Ning, Munan and Yan, Yang and Cui, Jiaxi and Wang, HongFa and Pang, Yatian and Jiang, Wenhao and Zhang, Junwu and Li, Zongwei and others},
  journal={arXiv preprint arXiv:2310.01852},
  year={2023}
}

@article{fvd,
  title={Towards accurate generative models of video: A new metric \& challenges},
  author={Unterthiner, Thomas and Van Steenkiste, Sjoerd and Kurach, Karol and Marinier, Raphael and Michalski, Marcin and Gelly, Sylvain},
  journal={arXiv preprint arXiv:1812.01717},
  year={2018}
}

@article{lin2023video,
  title={Video-llava: Learning united visual representation by alignment before projection},
  author={Lin, Bin and Ye, Yang and Zhu, Bin and Cui, Jiaxi and Ning, Munan and Jin, Peng and Yuan, Li},
  journal={arXiv preprint arXiv:2311.10122},
  year={2023}
}

@article{lin2024moe,
  title={Moe-llava: Mixture of experts for large vision-language models},
  author={Lin, Bin and Tang, Zhenyu and Ye, Yang and Cui, Jiaxi and Zhu, Bin and Jin, Peng and Huang, Jinfa and Zhang, Junwu and Pang, Yatian and Ning, Munan and others},
  journal={arXiv preprint arXiv:2401.15947},
  year={2024}
}

@inproceedings{li2025wf,
  title={Wf-vae: Enhancing video vae by wavelet-driven energy flow for latent video diffusion model},
  author={Li, Zongjian and Lin, Bin and Ye, Yang and Chen, Liuhan and Cheng, Xinhua and Yuan, Shenghai and Yuan, Li},
  booktitle={Proceedings of the Computer Vision and Pattern Recognition Conference},
  pages={17778--17788},
  year={2025}
}

@article{chen2024sharegpt4video,
  title={Sharegpt4video: Improving video understanding and generation with better captions},
  author={Chen, Lin and Wei, Xilin and Li, Jinsong and Dong, Xiaoyi and Zhang, Pan and Zang, Yuhang and Chen, Zehui and Duan, Haodong and Tang, Zhenyu and Yuan, Li and others},
  journal={Advances in Neural Information Processing Systems},
  volume={37},
  pages={19472--19495},
  year={2024}
}
\bibliographystyle{iclr2026_conference}
\newpage % 强制分页
\appendix
\section*{\centering
\textsc{\textbf{\method}: \textbf{F}ourier-Guided \textbf{La}tent \textbf{Sh}ifting Prevents Conditional Image Leakage in Image-to-Video Generation} \\
\Large Appendix
}

% comment for arxiv
% \section{The Use of Large Language Models (LLMs)}
% \label{app:using_of_llm}
% This work utilizes various LLMs, and their roles are as follows:
% \begin{itemize}
% \item {\bf Polishing tool}. The authors utilize the GPT-4o as a tool to polish and refine the writing.
% \item {\bf Recaptioning for text-image pairs in Vbench-I2V}. Since the texts of the text-video pairs in the training set consist solely of long texts, and the prompts used in Vbench-I2V~\citep{vbench,vbench2} are short, we use the gpt-4.1-2025-04-14 API to perform recaptioning on each image in Vbench-I2V based on its provided short prompt.
% \item {\bf Captioning for videos in training data}. Following Open-Sora Plan~\citep{open_sora_plan}, we annotate our data with QWen2-VL-7B~\citep{qwen2} using the same prompt. 
% \end{itemize}

\section{Recaptioning for Text-image Pairs in Vbench-I2V}
\label{app:recaptioning_vbench_i2v}
\method is initialized from the Wan2.1-T2V-1.3B weights. Since Wan2.1 is trained on text-video pairs with long captions, and our training set also consists of such pairs, we perform recaptioning for the text-image pairs of Vbench-I2V to accurately evaluate model performance.

We use the GPT-4.1-2025-04-14 API for recaptioning, with the following prompt design:

\begin{minipage}{\textwidth}
\ttfamily
You are a professional text editor, skilled at optimizing video descriptions. You will be given an image and a text description of the video's content, starting with the input image.

Please polish the input description to make it more vivid, concise, and expressive, while preserving the original meaning.

Please limit the polished description to between 100-150 words. The new prompt directly describes the specific content without words such as video, image, or picture.

description: \{caption\}.
\end{minipage}

\begin{figure}[t]
    \centering
    \includegraphics[width=1.0\linewidth]{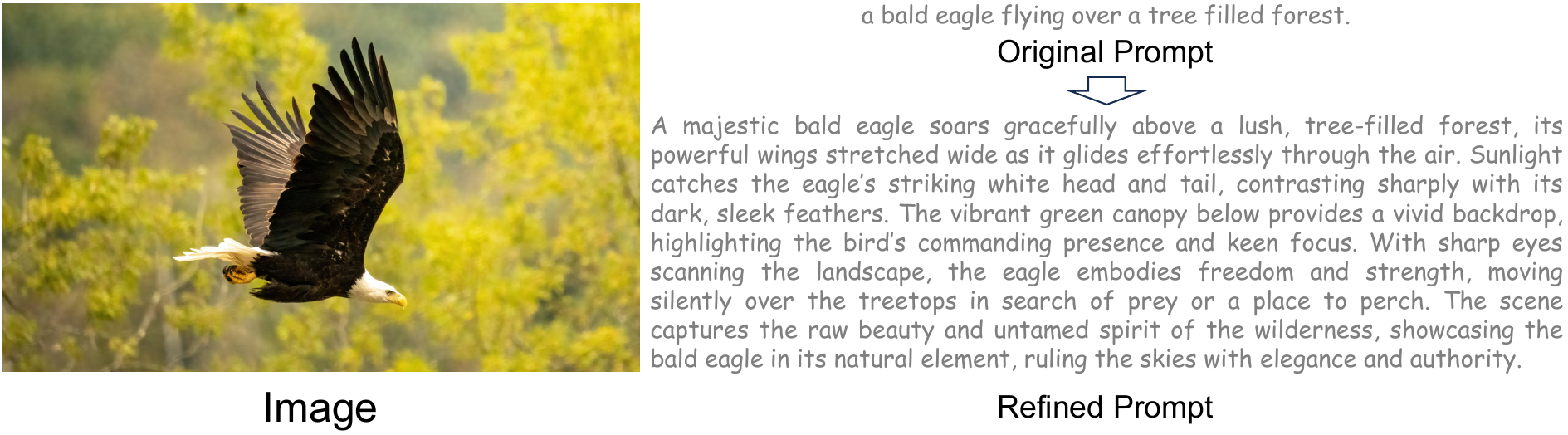}
    \caption{{\bf Racaptioning for text-image pairs in Vbench-I2V}. With recaptioning, the refined prompt now includes more detailed descriptions, aligning better with the distribution of the training set, which enhances the inference performance of the model to a more realistic result.}
    \label{fig:appendix_recaption}
\end{figure}

As shown in Fig.~\ref{fig:appendix_recaption}, the original short prompt is refined into a long prompt, adding more detailed descriptions while maintaining the original meaning.

\section{High-frequency Magnitude Features in Fourier Guidance}
\label{app:fourier_guidance}
In Sec.~\ref{method:fourier_guidance}, we discuss using the high-frequency magnitude features extracted through the Fourier transform as guidance to accelerate convergence. The complete derivation is as follows:

Let $\mathbf{FFT}(\cdot)$ denote the Fourier Transform and $\mathbf{iFFT}(\cdot)$ denote the inverse Fourier Transform. We first apply $\mathbf{FFT}(\cdot)$ to the conditional image latents $\boldsymbol{s}$ to obtain the frequency spectrum of $\boldsymbol{s}$:
\begin{equation}
    \begin{aligned}
        \mathbf{FFT}(\boldsymbol{s}) &\triangleq \hat{\boldsymbol{s}}_c^{\mathrm{freq}}(u,v) \\
        &=\sum_{h=0}^{H-1}\sum_{w=0}^{W-1}\boldsymbol{s}_c(h,w)\exp\left(-2\pi i\left(\frac{uh}{H}+\frac{vw}{W}\right)\right),
    \end{aligned}
\end{equation}
where $\boldsymbol{s}_c(h,w)$ is the value in the c-th channel in the spatial domain and $\hat{\boldsymbol{s}}_c^{\mathrm{freq}}(u,v)$ is the complex frequency data in the frequency domain. We remove the phase information from the frequency spectrum and retain only the magnitude, resulting in the magnitude map $\boldsymbol{M}_c(u,v)$:
\begin{equation}
    \boldsymbol{M}_c(u,v)=|\hat{\boldsymbol{s}}_c^{\mathrm{freq}}(u,v)|=\sqrt{\Re(\hat{\boldsymbol{s}}_c^{\mathrm{freq}}(u,v))^2+\Im(\hat{\boldsymbol{s}}_c^{\mathrm{freq}}(u,v))^2},
\end{equation}
where $\Re(\hat{\boldsymbol{s}}_c^{\mathrm{freq}}(u,v))$ and $\Im(\hat{\boldsymbol{s}}_c^{\mathrm{freq}}(u,v))$ represent the real and imaginary parts of the complex frequency, respectively. Then, a specified cutoff frequency $\mathrm{cutoff}\_\mathrm{freq}$ is defined as:
\begin{equation}
\mathrm{cutoff}\_\mathrm{freq}=\min \{r\mid \frac{\sum_{r(u,v)\le r}{\boldsymbol{M}(u,v)}}{\sum_{u,v}{\boldsymbol{M}(u,v)}}\ge p\},
\end{equation}
where $p$ is the cutoff frequency percentile, representing the percentile of low-frequency energy. In addition, $r(u,v)$ represents the radius, which can be calculated using the following formula:
\begin{equation}
    r(u,v)=\sqrt{(u-u_0)^2+(v-v_0)^2},
\end{equation}
where $(u_0,v_0)$ is the center of the frequency plane. To implement high-frequency information extraction, we define the frequency masks:
\begin{equation}
    \begin{aligned}
     & \mathbf{Mask}^{\mathrm{low}}(u,v)=
    \begin{cases}
    1, & \mathrm{if~}r(u,v)<\mathrm{cutoff}\_\mathrm{freq}, \\
    0, & \mathrm{otherwise}, 
    \end{cases} \\
     & \mathbf{Mask}^{\mathrm{high}}(u,v)=1-\mathbf{Mask}^{\mathrm{low}}(u,v).
    \end{aligned}
\end{equation}  
After performing filtering in the frequency domain using the frequency mask $\mathbf{Mask}^{\mathrm{high}}$, we can apply the inverse Fourier transform $\mathbf{iFFT}(\cdot)$ to obtain the high-frequency component in the spatial domain $\boldsymbol{s}_c^\mathrm{high}$:
\begin{equation}
\boldsymbol{s}_c^\mathrm{high}=\mathbf{iFFT}\left(\hat{\boldsymbol{s}}_c^\mathrm{freq}\cdot \mathbf{Mask}^{\mathrm{high}}\right).
\end{equation}
After performing the inverse Fourier Transform to obtain the high-frequency component $\boldsymbol{s}_c^\mathrm{high}$, we perform magnitude extraction to obtain the final magnitude of the high-frequency component $\boldsymbol{M}_c^{\mathrm{high}}$:
\begin{equation}
\boldsymbol{M}_c^{\mathrm{high}}=|\boldsymbol{s}_c^{\mathrm{high}}|.
\end{equation}
The $\boldsymbol{s}_{\text{high}}$ in each channel from Sec.~\ref{method:fourier_guidance} corresponds to $\boldsymbol{M}_c^{\mathrm{high}}$ here.

\begin{figure}[t]
    \centering
    \includegraphics[width=\linewidth]{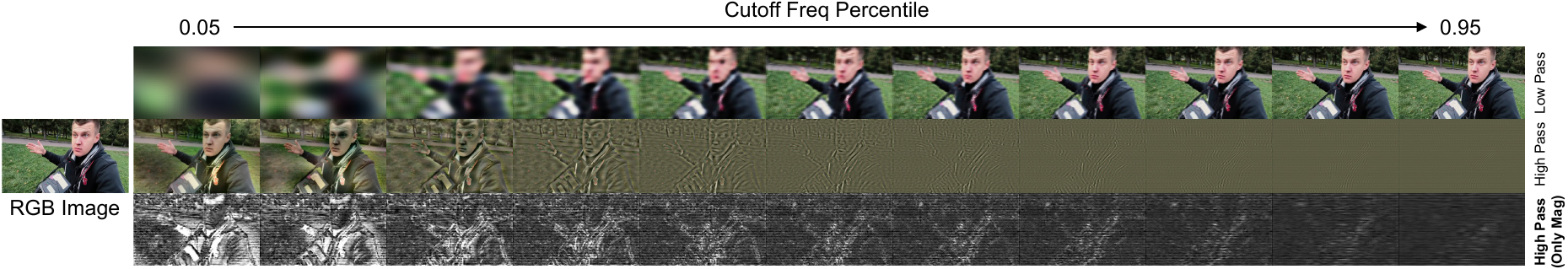}
    \caption{{\bf Information extracted by the Fourier Transform}. After performing the Fourier Transform in the latent space and decoding features to the pixel space, we observe that as the cutoff frequency percentile increases, the high-frequency information in the latents diminishes. We extract only the magnitude of the high-frequency information, ensuring that the original high-frequency information cannot be restored while still providing guidance.}
    \label{fig:appendix_fourier_frequency}
\end{figure}

As shown in Fig.~\ref{fig:appendix_fourier_frequency}, in the latent space, the amount of high-frequency information decreases as the cutoff frequency percentile increases, similar to the behavior in pixel space. Therefore, for a conditional image, we apply the above operation in the latent space to save computational resources. If we use the original extracted high-frequency features, the resulting features still contain information such as color, which can be easily shortcut. By retaining only the magnitude, we preserve the relative strength of the signal, thus emphasizing the role of guidance without a shortcut.

\section{Pseudo-code implementation of different I2V paradigms}
\label{app:code_implementation}
\begin{algorithm}
\caption{Sampling process for Repeating Concat paradigm}
\label{alg:appendix_repeating_concat}
    \begin{algorithmic}[1]
      \Require Denoiser \(\boldsymbol{v}_\theta\), VAE encoder \(\mathcal{E}\), VAE decoder \(\mathcal{D}\), input conditional image \(\boldsymbol{S}\), prompt \(\boldsymbol{y}\), guidance \(w\), total inference steps \(N\)
      \State \(\boldsymbol{s} \gets \mathcal{E}(\boldsymbol{S})\) \State \(\boldsymbol{s} \gets \texttt{Repeat}(\boldsymbol{s})\)
      \State \(\boldsymbol{z} \sim \mathcal{N}(\boldsymbol{0}, \boldsymbol{I})\) 
      \For{\(i = 0\) to \(N-1\)}                 
        \State \(t \gets \tfrac{N-i}{N}\)
        \State \(\hat{\boldsymbol{z}} \gets \texttt{Concat}(\boldsymbol{s}, \boldsymbol{z}) \)
        \State \(\boldsymbol{v} \gets \boldsymbol{v}_\theta (\hat{\boldsymbol{z}},t,\varnothing) + w[\boldsymbol{v}_\theta (\hat{\boldsymbol{z}},t,\boldsymbol{y}) - 
        \boldsymbol{v}_\theta (\hat{\boldsymbol{z}},t,\varnothing)]\)
        \State \(\boldsymbol{z} \gets \texttt{SolverStep} (\boldsymbol{z},\boldsymbol{v},t)\)
      \EndFor
      \State \Return \(\mathcal{D}(\boldsymbol{z})\)
    \end{algorithmic}
\end{algorithm}

\begin{algorithm}[t!]
\caption{Sampling process for Repeating Concat and Adding Noise paradigm}
\label{alg:appendix_repeating_concat_and_adding_noise}
\begin{algorithmic}[1]
  \Require Denoiser \(\boldsymbol{v}_\theta\), VAE encoder \(\mathcal{E}\), VAE decoder \(\mathcal{D}\), input conditional image \(\boldsymbol{S}\), prompt \(\boldsymbol{y}\), guidance \(w\), total inference steps \(N\), mean of adding noise \(\boldsymbol{\mu}\), variance of adding noise \(\boldsymbol{\sigma^2}\)
  \State \(\boldsymbol{\epsilon} \sim \mathcal{N}(\boldsymbol{\mu}, \boldsymbol{\sigma^2})\) 
  \State \(\boldsymbol{S} \gets \texttt{Add\_Noise}(\boldsymbol{S},\boldsymbol{\epsilon})\)
  \State \(\boldsymbol{s} \gets \mathcal{E}(\boldsymbol{S})\)
  \State \(\boldsymbol{s} \gets \texttt{Repeat}(\boldsymbol{s})\)
  \State \(\boldsymbol{z} \sim \mathcal{N}(\boldsymbol{0}, \boldsymbol{I})\) 
  \For{\(i = 0\) to \(N-1\)}                 
    \State \(t \gets \tfrac{N-i}{N}\)
    \State \(\hat{\boldsymbol{z}} \gets \texttt{Concat}(\boldsymbol{s}, \boldsymbol{z})\)
    \State \(\boldsymbol{v} \gets \boldsymbol{v}_\theta (\hat{\boldsymbol{z}},t,\varnothing) + w[\boldsymbol{v}_\theta (\hat{\boldsymbol{z}},t,\boldsymbol{y}) - 
    \boldsymbol{v}_\theta (\hat{\boldsymbol{z}},t,\varnothing)]\)
    \State \(\boldsymbol{z} \gets \texttt{SolverStep} (\boldsymbol{z},\boldsymbol{v},t)\)
  \EndFor
  \State \Return \(\mathcal{D}(\boldsymbol{z})\)
\end{algorithmic}
\end{algorithm}
\begin{algorithm}[t!]
\caption{Sampling process for Zero-Padding Concat paradigm}
\label{alg:appendix_zero_padding_concat}
\begin{algorithmic}[1]
  \Require Denoiser \(\boldsymbol{v}_\theta\), VAE encoder \(\mathcal{E}\), VAE decoder \(\mathcal{D}\), input conditional image \(\boldsymbol{S}\), prompt \(\boldsymbol{y}\), guidance \(w\), total inference steps \(N\)
  \State \(\boldsymbol{s} \gets \mathcal{E}(\boldsymbol{S})\) \State \(\boldsymbol{s} \gets \texttt{Pad\_Zeros}(\boldsymbol{s})\)
  \State \(\boldsymbol{z} \sim \mathcal{N}(\boldsymbol{0}, \boldsymbol{I})\) 
  \For{\(i = 0\) to \(N-1\)}                 
    \State \(t \gets \tfrac{N-i}{N}\)
    \State \(\hat{\boldsymbol{z}} \gets \texttt{Concat}(\boldsymbol{s}, \boldsymbol{z})\)
    \State \(\boldsymbol{v} \gets \boldsymbol{v}_\theta (\hat{\boldsymbol{z}},t,\varnothing) + w[\boldsymbol{v}_\theta (\hat{\boldsymbol{z}},t,\boldsymbol{y}) - 
    \boldsymbol{v}_\theta (\hat{\boldsymbol{z}},t,\varnothing)]\)
    \State \(\boldsymbol{z} \gets \texttt{SolverStep} (\boldsymbol{z},\boldsymbol{v},t)\)
  \EndFor
  \State \Return \(\mathcal{D}(\boldsymbol{z})\)
\end{algorithmic}
\end{algorithm}
\begin{algorithm}[t!]
\caption{Sampling process for Zero-Padding Concat and Adding Noise paradigm}
\label{alg:appendix_zero_padding_concat_and_adding_noise}
\begin{algorithmic}[1]
  \Require Denoiser \(\boldsymbol{v}_\theta\), VAE encoder \(\mathcal{E}\), VAE decoder \(\mathcal{D}\), input conditional image \(\boldsymbol{S}\), prompt \(\boldsymbol{y}\), guidance \(w\), total inference steps \(N\), mean of adding noise \(\boldsymbol{\mu}\), variance of adding noise \(\boldsymbol{\sigma^2}\)
  \State \(\boldsymbol{\epsilon} \sim \mathcal{N}(\boldsymbol{\mu}, \boldsymbol{\sigma^2})\) 
  \State \(\boldsymbol{S} \gets \texttt{Add\_Noise}(\boldsymbol{S},\boldsymbol{\epsilon})\)
  \State \(\boldsymbol{s} \gets \mathcal{E}(\boldsymbol{S})\)
  \State \(\boldsymbol{s} \gets \texttt{Pad\_Zeros}(\boldsymbol{s})\)
  \State \(\boldsymbol{z} \sim \mathcal{N}(\boldsymbol{0}, \boldsymbol{I})\) 
  \For{\(i = 0\) to \(N-1\)}                 
    \State \(t \gets \tfrac{N-i}{N}\)
    \State \(\hat{\boldsymbol{z}} \gets \texttt{Concat}(\boldsymbol{s}, \boldsymbol{z})\)
    \State \(\boldsymbol{v} \gets \boldsymbol{v}_\theta (\hat{\boldsymbol{z}},t,\varnothing) + w[\boldsymbol{v}_\theta (\hat{\boldsymbol{z}},t,\boldsymbol{y}) - 
    \boldsymbol{v}_\theta (\hat{\boldsymbol{z}},t,\varnothing)]\)
    \State \(\boldsymbol{z} \gets \texttt{SolverStep} (\boldsymbol{z},\boldsymbol{v},t)\)
  \EndFor
  \State \Return \(\mathcal{D}(\boldsymbol{z})\)
\end{algorithmic}
\end{algorithm}

\begin{algorithm}[t!]
\caption{Sampling process for Inpainting paradigm}
\label{alg:appendix_inpainting}
\begin{algorithmic}[1]
  \Require Denoiser \(\boldsymbol{v}_\theta\), VAE encoder \(\mathcal{E}\), VAE decoder \(\mathcal{D}\), input conditional image \(\boldsymbol{S}\), prompt \(\boldsymbol{y}\), guidance \(w\), total inference steps \(N\)
  \State \(\boldsymbol{S} \gets \texttt{Pad\_Zeros}(\boldsymbol{S}) \)
  \State \(\mathbf{M} \gets \texttt{Generate\_Mask}(\boldsymbol{S})\)
  \State \(\mathbf{m} \gets \texttt{Downsample\_And\_Rearrange}(\boldsymbol{\mathbf{M}})\)
  \State \(\boldsymbol{s} \gets \mathcal{E}(\boldsymbol{S})\)
  \State \(\boldsymbol{z} \sim \mathcal{N}(\boldsymbol{0}, \boldsymbol{I})\) 
  \For{\(i = 0\) to \(N-1\)}                 
    \State \(t \gets \tfrac{N-i}{N}\)
    \State \(\hat{\boldsymbol{z}} \gets \texttt{Concat}(\mathbf{m},\boldsymbol{s}, \boldsymbol{z})\)
    \State \(\boldsymbol{v} \gets \boldsymbol{v}_\theta (\hat{\boldsymbol{z}},t,\varnothing) + w[\boldsymbol{v}_\theta (\hat{\boldsymbol{z}},t,\boldsymbol{y}) - 
    \boldsymbol{v}_\theta (\hat{\boldsymbol{z}},t,\varnothing)]\)
    \State \(\boldsymbol{z} \gets \texttt{SolverStep} (\boldsymbol{z},\boldsymbol{v},t)\)
  \EndFor
  \State \Return \(\mathcal{D}(\boldsymbol{z})\)
\end{algorithmic}
\end{algorithm}

\begin{algorithm}[t!]
\caption{Sampling process for Inpainting and Adding Noise paradigm}
\label{alg:appendix_inpainting_and_adding_noise}
\begin{algorithmic}[1]
  \Require Denoiser \(\boldsymbol{v}_\theta\), VAE encoder \(\mathcal{E}\), VAE decoder \(\mathcal{D}\), input conditional image \(\boldsymbol{S}\), prompt \(\boldsymbol{y}\), guidance \(w\), total inference steps \(N\), mean of adding noise \(\boldsymbol{\mu}\), variance of adding noise \(\boldsymbol{\sigma^2}\)
  \State \(\boldsymbol{\epsilon} \sim \mathcal{N}(\boldsymbol{\mu}, \boldsymbol{\sigma^2})\) 
  \State \(\boldsymbol{S} \gets \texttt{Add\_Noise}(\boldsymbol{S},\boldsymbol{\epsilon})\)
  \State \(\boldsymbol{S} \gets \texttt{Pad\_Zeros}(\boldsymbol{S}) \)
  \State \(\mathbf{M} \gets \texttt{Generate\_Mask}(\boldsymbol{S})\)
  \State \(\mathbf{m} \gets \texttt{Downsample\_And\_Rearrange}(\boldsymbol{\mathbf{M}})\)
  \State \(\boldsymbol{s} \gets \mathcal{E}(\boldsymbol{S})\)
  \State \(\boldsymbol{z} \sim \mathcal{N}(\boldsymbol{0}, \boldsymbol{I})\) 
  \For{\(i = 0\) to \(N-1\)}                 
    \State \(t \gets \tfrac{N-i}{N}\)
    \State \(\hat{\boldsymbol{z}} \gets \texttt{Concat}(\mathbf{m},\boldsymbol{s}, \boldsymbol{z})\)
    \State \(\boldsymbol{v} \gets \boldsymbol{v}_\theta (\hat{\boldsymbol{z}},t,\varnothing) + w[\boldsymbol{v}_\theta (\hat{\boldsymbol{z}},t,\boldsymbol{y}) - 
    \boldsymbol{v}_\theta (\hat{\boldsymbol{z}},t,\varnothing)]\)
    \State \(\boldsymbol{z} \gets \texttt{SolverStep} (\boldsymbol{z},\boldsymbol{v},t)\)
  \EndFor
  \State \Return \(\mathcal{D}(\boldsymbol{z})\)
\end{algorithmic}
\end{algorithm}

\begin{algorithm}[t!]
\caption{Sampling process for \method paradigm}
\label{alg:appendix_flashi2v}
\begin{algorithmic}[1]
  \Require Denoiser \(\boldsymbol{v}_\theta\), learnable projection \(\boldsymbol{\phi}\), VAE encoder \(\mathcal{E}\), VAE decoder \(\mathcal{D}\), input conditional image \(\boldsymbol{S}\), prompt \(\boldsymbol{y}\), guidance \(w\), total inference steps \(N\).
  \State \(\boldsymbol{s} \gets \mathcal{E}(\boldsymbol{S})\)
  \State \(\boldsymbol{s}_{\text{high}} \gets \texttt{Fourier\_Filter}(\boldsymbol{s})\)
  \State \(\boldsymbol{z} \sim \mathcal{N}(\boldsymbol{0}, \boldsymbol{I})\) 
  \For{\(i = 0\) to \(N-1\)}                 
    \State \(t \gets \tfrac{N-i}{N}\)
    \State \(\hat{\boldsymbol{z}} \gets \texttt{Concat}(\boldsymbol{s}_{\text{high}}, \boldsymbol{z}-\boldsymbol{\phi}(\boldsymbol{s}))\)
    \State \(\boldsymbol{v} \gets \boldsymbol{v}_\theta (\hat{\boldsymbol{z}},t,\varnothing) + w[\boldsymbol{v}_\theta (\hat{\boldsymbol{z}_t},t,\boldsymbol{y}) - 
    \boldsymbol{v}_\theta (\hat{\boldsymbol{z}_t},t,\varnothing)]\)
    \State \(\boldsymbol{z} \gets \texttt{SolverStep} (\boldsymbol{z},\boldsymbol{v},t)\)
  \EndFor
  \State \Return \(\mathcal{D}(\boldsymbol{z})\)
\end{algorithmic}
\end{algorithm}
In Sec.~\ref{exp:ablation_study}, we compare the chunk-wise FVD of \method with existing I2V paradigms. The pseudo-code implementations for the sampling process of various paradigms are as follows.

{\bf Repeating Concat}. As illustrated in Algorithm~\ref{alg:appendix_repeating_concat}, this paradigm first repeats the conditional image latents along the temporal dimension, and then concatenates the repeated condition with noisy latents along the channel dimension to achieve I2V.

{\bf Repeating Concat and Adding Noise}. As shown in Algorithm~\ref{alg:appendix_repeating_concat_and_adding_noise}, compared to the Repeating Concat paradigm, this approach adds a small amount of noise to the conditional images. The intensity of the noise is not strong enough to disrupt most of the information in the conditional images, but it can enhance the generalization of the model. This paradigm is used in SVD~\citep{stable_video_diffusion}.

{\bf Zero-Padding Concat}. As shown in Algorithm~\ref{alg:appendix_zero_padding_concat}, this paradigm pads the conditional image latents $\boldsymbol{s}$ with zeros along the temporal dimension to match the shape of noisy latents, then concatenates them with noisy latents along the channel dimension.

{\bf Zero-Padding Concat and Adding Noise}. As shown in Algorithm~\ref{alg:appendix_zero_padding_concat_and_adding_noise}, compared to the Zero-padding Concat paradigm, this approach adds a small amount of noise to the conditional images. This paradigm is used in CogVideoX~\citep{cogvideox}.

{\bf Inpainting}. As shown in Algorithm~\ref{alg:appendix_inpainting}, this paradigm treats I2V as a temporal completion task. For the conditional image $\boldsymbol{S}$, the approach pads $\boldsymbol{S}$ with zeros along the temporal dimension to align with the video shape. After encoding with the VAE encoder, $\boldsymbol{s}$ is obtained. Additionally, a mask is generated based on $\boldsymbol{S}$ to identify frames with information, which is then downsampled and rearranged to align with the frame numbers, height, and width of latents, resulting in $\mathbf{m}$. Both $\mathbf{m}$ and $\boldsymbol{s}$ are concatenated with noisy latents along the channel dimension as input to the denoiser. This paradigm is used in Open-Sora Plan~\citep{open_sora_plan} and Wan2.1~\citep{wan}.

{\bf Inpainting and Adding Noise}. Algorithm~\ref{alg:appendix_inpainting_and_adding_noise} shows that compared to the Inpainting paradigm, this paradigm first adds a small amount of noise to the conditional image, followed by temporal inpainting.

{\bf FlashI2V (Ours)}. As shown in Algorithm~\ref{alg:appendix_flashi2v}, FlashI2V modifies only the input of the denoiser compared to other paradigms. First, the conditional image latents are encoded through a learnable projection to obtain $\boldsymbol{\phi}(\boldsymbol{s})$, which acts as the shifting for the noisy latents $\boldsymbol{z}_t$. Additionally, high-frequency magnitude features $\boldsymbol{s}_{\text{high}}$ are extracted through the Fourier Transform. $\boldsymbol{s}_{\text{high}}$ and $\boldsymbol{z}_t - \boldsymbol{\phi}(\boldsymbol{s})$ are concatenated together as inputs to the denoiser. According to the derivation in Sec.~\ref{method:latent_shifting}, we can deduce that the resulting $\boldsymbol{v}$ is the velocity field conditioned on the input image.
\vspace{-0.5em}
\section{Further Experiments on Fourier Cutoff Frequency}
\label{app:further_experiments_on_fourier}
In Sec.~\ref{exp:analysis}, we point out that the results obtained at lower cutoff frequencies exhibit higher fidelity, and the details in the video are more consistently preserved at inference. During training, we sample cutoff frequency percentiles from $\mathcal{U}[0.05, 0.95]$. In this section, we test the effect of using lower cutoff frequencies during training.

\begin{figure}[t]
    \centering
    \begin{subfigure}[b]{0.24\textwidth}
        \centering
        \includegraphics[width=\textwidth]{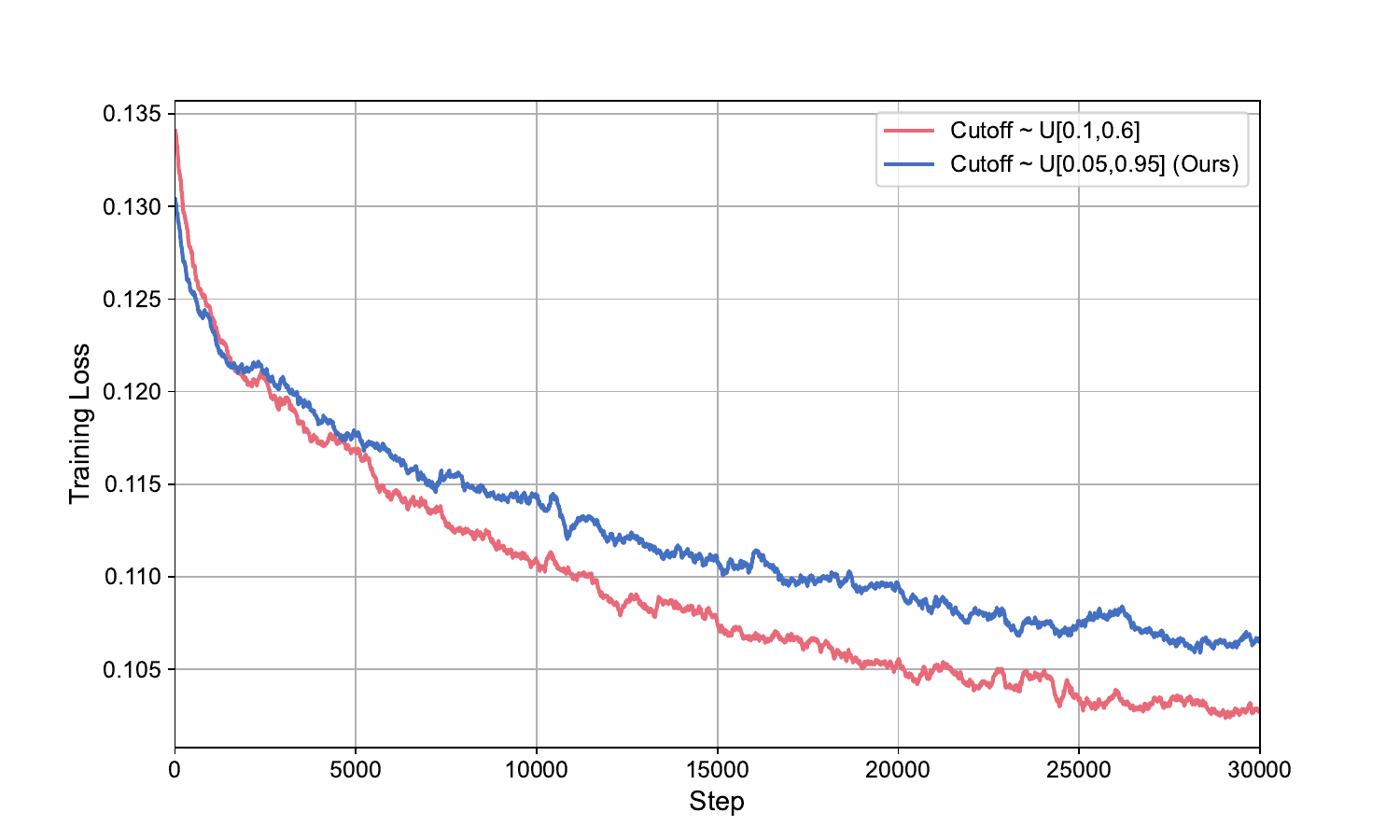}
        \caption{Training Loss}
        \label{fig:appendix_abs_change_training_loss}
    \end{subfigure}
    \begin{subfigure}[b]{0.74\textwidth}
        \centering
        \includegraphics[width=\textwidth]{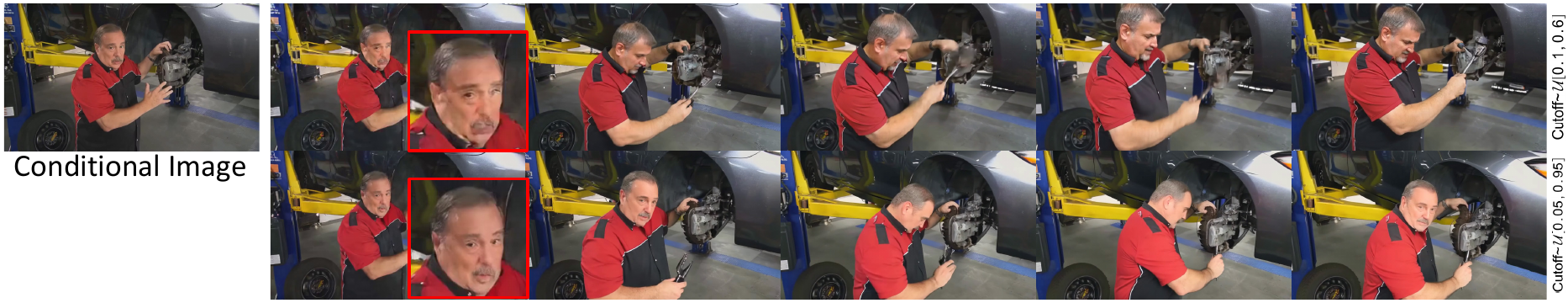}
        \caption{Qualitative Results}
        \label{fig:appendix_abs_change_qualitative}
    \end{subfigure}
    \caption{{\bf The impact of different cutoff frequency percentiles during training}. (a) Using generally lower cutoff frequency percentiles results in a lower training loss. (b) Training with lower cutoff frequency percentiles leads to worse fidelity in inference. This suggests that during training, it is important to reduce the injection of high-frequency information appropriately, as an excessive input of high-frequency features can negatively affect performance.}
    \label{fig:appendix_abs_change}
\end{figure}

As shown in Fig.~\ref{fig:appendix_abs_change}, using cutoff frequency percentiles sampled from $\mathcal{U}[0.1, 0.6]$ compared to $\mathcal{U}[0.05, 0.95]$ results in a higher probability of encountering lower cutoff frequencies, leading to a lower training loss. However, the fidelity of details in the sampling results decreases. This is because if only lower cutoff frequencies are encountered during training, the training process is dominated by Fourier guidance, and the learnable projection in latent shifting cannot be fully trained.
\vspace{-0.5em}
\section{Data Collection and Processing}
\label{app:data_collection_and_processing}
Our training dataset incorporates some internal data and open-source data from Panda-70M~\citep{panda70m} and VIDAL~\citep{languagebind}. We also include videos from CC0-licensed websites such as Mixkit, Pexels, and Pixabay. In addition, our data collection and processing pipeline follows the Open-Sora Plan~\citep{open_sora_plan}. 

\section{Models and Codes Used in Experiments}
\label{app:models_and_codes_links}
As shown in Tab.~\ref{tab:appendix_model_source}, we provide model links to all Vbench-I2V metrics used in Sec.~\ref{exp:comparisons} and Sec.~\ref{exp:ablation_study}. The table also includes the link to the FVD implementation, as well as links to the CogVideoX1.5-5B-I2V and Wan2.1-I2V-14B-480P models and codes, ensuring the reproducibility of our results.
\begin{table}[t!]
    \centering
    \caption{The source links of models and codes used in our experiments.}
    \resizebox{\textwidth}{!}{%
    \setlength\tabcolsep{3pt}
    \begin{tabular}{lc}
        \toprule
        Metric or Model & Source Link \\
        \midrule
        FVD & \url{https://github.com/JunyaoHu/common_metrics_on_video_quality} \\
        Subject Consistency & \url{https://dl.fbaipublicfiles.com/dino/dino_vitbase16_pretrain/dino_vitbase16_pretrain.pth} \\
        Background Consistency & \url{https://openaipublic.azureedge.net/clip/models/40d365715913c9da98579312b702a82c18be219cc2a73407c4526f58eba950af/ViT-B-32.pt} \\
        Motion Smoothness & \url{https://huggingface.co/lalala125/AMT/resolve/main/amt-s.pth} \\
        Dynamic Degree & \url{https://dl.dropboxusercontent.com/s/4j4z58wuv8o0mfz/models.zip} \\
        Aesthetic Quality & \url{https://huggingface.co/sentence-transformers/clip-ViT-L-14} \\
        Imaging Quality & \url{https://github.com/chaofengc/IQA-PyTorch/releases/download/v0.1-weights/musiq_spaq_ckpt-358bb6af.pth} \\
        I2V Subject Consistency & \url{https://dl.fbaipublicfiles.com/dino/dino_vitbase16_pretrain/dino_vitbase16_pretrain.pth} \\
        I2V Background Consistency & \url{https://github.com/ssundaram21/dreamsim/releases/download/v0.2.0-checkpoints/dreamsim_ensemble_checkpoint.zip} \\
        \midrule
        CogVideoX1.5-5B-I2V & \makecell{\url{https://github.com/zai-org/CogVideo}\\ \url{https://huggingface.co/zai-org/CogVideoX1.5-5B-I2V}} \\
        Wan2.1-I2V-14B-480P & \makecell{\url{https://github.com/Wan-Video/Wan2.1}\\ \url{https://huggingface.co/Wan-AI/Wan2.1-I2V-14B-480P}} \\
        \bottomrule
    \end{tabular}
    }
    \label{tab:appendix_model_source}
    \vspace{-1em}
\end{table}
\begin{table}[t!]
    \centering
    \samepage
    \caption{{\bf FVD values across different I2V paradigms} on training set and validation set. The quantitative results show that only \method exhibits the same FVD variation pattern on both in-domain and out-of-domain data.}
    \begin{subtable}{0.48\textwidth}
        \caption{FVD on training set}
        \resizebox{\textwidth}{!}{
            \begin{tabular}{lccccc}
            \toprule
            Method & T[0:12]$\downarrow$ & T[12:24]$\downarrow$ & T[24:36]$\downarrow$ & T[36:48]$\downarrow$ & Overall$\downarrow$ \\
            \midrule
            Repeating Concat & 99.27 & 133.51 & 147.12 & 145.24 & 122.00  \\
            Repeating Concat and Add Noise & 89.30 & 127.12 & 142.62 & 139.85 & 119.14  \\
            Zero-padding Concat & 103.37 & 144.44 & 167.21 & 166.17 & 124.50  \\
            Zero-padding Concat and Add Noise & 98.86 & 137.92 & 161.54 & 157.28 & 137.03  \\
            Inpainting & 104.37 & 140.39 & 155.19 & 154.20 & 131.84  \\
            Inpainting and Add Noise & 116.94 & 154.23 & 170.24 & 173.45 & 148.86  \\
            \midrule
            \rowcolor[HTML]{dae7ed} {\bf \method} & {\bf 81.42} & {\bf 109.59} & {\bf 116.93} & {\bf 111.25} & {\bf 103.39} \\
            \bottomrule
        \end{tabular}
        }
    \end{subtable}
    \begin{subtable}{0.48\textwidth}
        \caption{FVD on validation set}
        \resizebox{\textwidth}{!}{
            \begin{tabular}{lccccc}
            \toprule
            Method & T[0:12]$\downarrow$ & T[12:24]$\downarrow$ & T[24:36]$\downarrow$ & T[36:48]$\downarrow$ & Overall$\downarrow$ \\
            \midrule
            Repeating Concat & 174.86 & 187.03 & 173.08 & 172.70 & 148.92  \\
            Repeating Concat and Add Noise & 163.29 & 167.10 & 150.86 & 154.84 & 143.43  \\
            Zero-padding Concat & 174.74 & 182.72 & 185.05 & 183.93 & 139.28  \\
            Zero-padding Concat and Add Noise & 157.28 & 165.44 & 158.16 & 159.13 & 127.68  \\
            Inpainting & 197.13 & 198.75 & 198.93 & 185.20 & 151.76  \\
            Inpainting and Add Noise & 251.31 & 232.13 & 210.95 & 190.42 & 156.96  \\
            \midrule
            \rowcolor[HTML]{dae7ed} {\bf \method} & {\bf 95.29} & {\bf 127.64} & {\bf 139.70} & {\bf 146.26} & {\bf 104.21} \\
            \bottomrule
        \end{tabular}
        }
    \end{subtable}
    \vspace{-1em}
    \label{tab:appendix_various_methods_fvd_values}
\end{table}
\begin{table}[t!]
    \samepage
    \centering
    \caption{{\bf FVD values across ablation experiments} of various modules in FlashI2V on training set and validation set. The quantitative results show that latent shifting with Fourier guidance results in the best generalization and performance on out-of-domain data.}
    \begin{subtable}{0.48\textwidth}
        \caption{FVD on training set}
        \resizebox{\textwidth}{!}{
            \begin{tabular}{lccccc}
            \toprule
            Method & T[0:12]$\downarrow$ & T[12:24]$\downarrow$ & T[24:36]$\downarrow$ & T[36:48]$\downarrow$ & Overall$\downarrow$ \\
            \midrule
            Only Latent Shifting & 86.84 & 118.82 & 126.38 & 123.13 & 111.35  \\
            Only Fourier Guidance & 159.14 & 198.89 & 210.23 & 206.68 & 158.47  \\
            \midrule
            \rowcolor[HTML]{dae7ed} {\bf \method} & {\bf 81.42} & {\bf 109.59} & {\bf 116.93} & {\bf 111.25} & {\bf 103.39} \\
            \bottomrule
        \end{tabular}
        }
    \end{subtable}
    \begin{subtable}{0.48\textwidth}
        \caption{FVD on validation set}
        \resizebox{\textwidth}{!}{
            \begin{tabular}{lccccc}
            \toprule
            Method & T[0:12]$\downarrow$ & T[12:24]$\downarrow$ & T[24:36]$\downarrow$ & T[36:48]$\downarrow$ & Overall$\downarrow$ \\
            \midrule
            Only Latent Shifting & 107.56 & {\bf 127.31} & {\bf 136.82} & {\bf 141.06} & 113.83  \\
            Only Fourier Guidance & 211.66 & 202.10 & 193.59 & 191.78 & 159.46  \\
            \midrule
            \rowcolor[HTML]{dae7ed} {\bf \method} & {\bf 95.29} & 127.64 &  139.70 & 146.26 & {\bf 104.21} \\
            \bottomrule
        \end{tabular}
        }
    \end{subtable}
    \label{tab:appendix_ablation_fvd_values}
    \vspace{-1em}
\end{table}
\vspace{-1em}
\section{More Quantified Results}
\label{app:more_quantified_results}
\subsection{FVD Values Across Various I2V Paradigms}
\label{app:more_quantified_results_i2v_paradigms}
In Fig.~\ref{fig:experiment_ablation_study_overfitting}, to facilitate the observation of the FVD variation patterns, we present the chunk-wise FVD of different I2V paradigms in the form of a bar chart. In Tab.~\ref{tab:appendix_various_methods_fvd_values}, we provide the specific values for the chunk-wise FVD and overall FVD of these paradigms. It can be observed that, compared to other I2V paradigms, \method shows consistent chunk-wise FVD variation patterns on both in-domain and out-of-domain data, and it achieves superior FVD across all chunks and overall, proving excellent generalization and performance.

\subsection{FVD Values Across Ablation Experiments of Various Modules in FlashI2V}
\label{app:more_quantified_results_ablation}
 In Sec.~\ref{exp:ablation_study}, we present the training loss and qualitative performance for the ablation experiments of different modules in \method. Here, we provide the FVD values of the ablation experiments, as shown in Tab.~\ref{tab:appendix_ablation_fvd_values}. It can be observed that latent shifting with Fourier guidance achieved the best FVD performance, demonstrating the effectiveness of \method.

\section{More Visual Results}
\label{app:more_visual_results}
\begin{figure}[t]
    \centering
    \includegraphics[width=\linewidth]{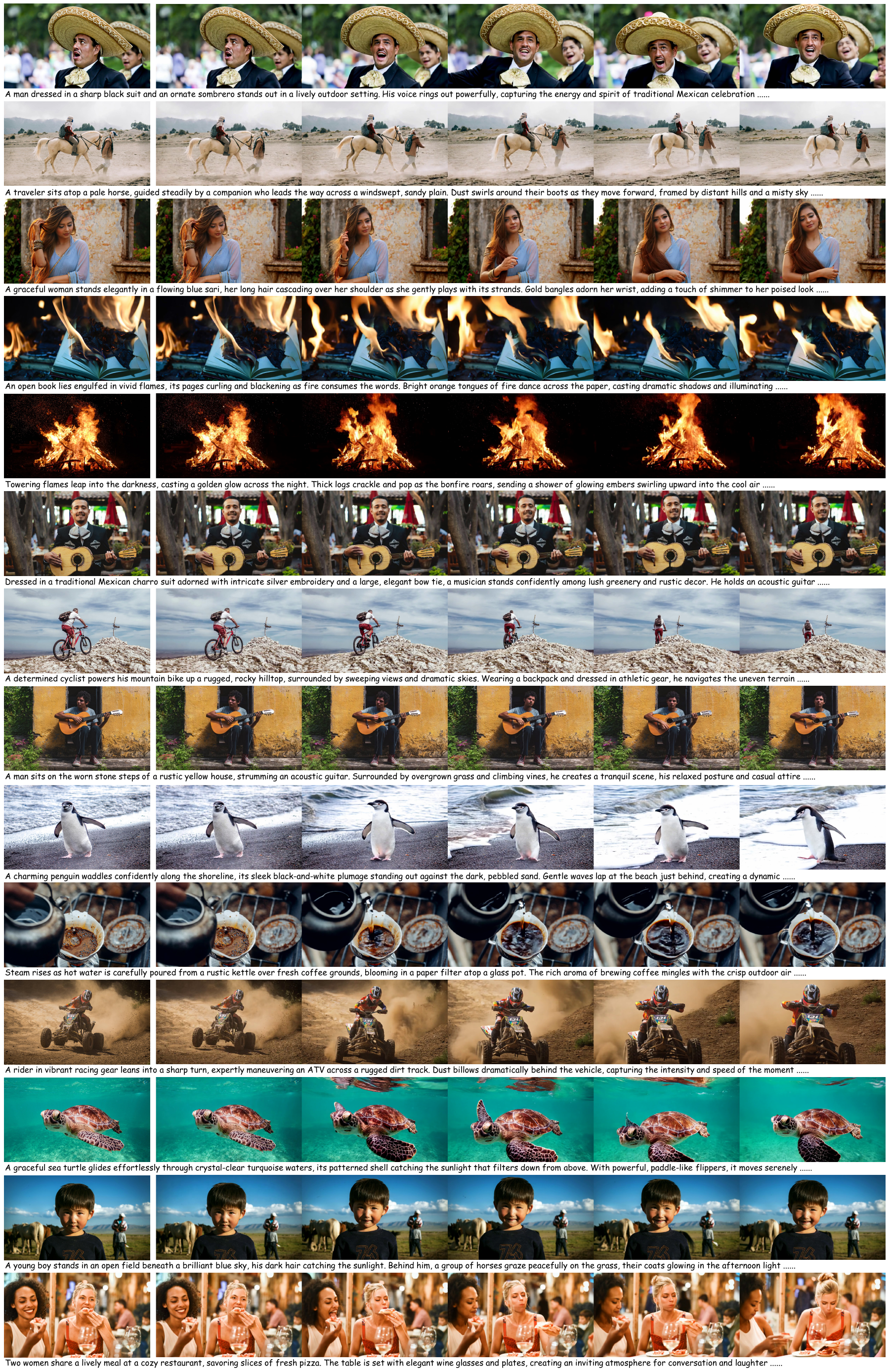}
    \caption{Visual results sampled from Vbench-I2V.}
    \label{fig:appendix_more_results}
\end{figure}

Fig.~\ref{fig:appendix_more_results} presents more inference results for \method, and the video version of the results can be found on the project page.

\end{document}